\pdfoutput=1
\documentclass[letterpaper,twocolumn,10pt]{article}

\usepackage{usenix,epsfig,endnotes}
\usepackage{balance} %
\usepackage{fullpage}
\usepackage{listings}
\usepackage{times}
\usepackage{color}
\usepackage[T1]{fontenc}    %
\usepackage{hyperref}       %
\usepackage{url}            %
\usepackage{booktabs}       %
\usepackage{amsfonts}       %
\usepackage{nicefrac}       %
\usepackage{microtype}      %
\usepackage{xspace}

\usepackage{graphicx}
\usepackage{subcaption}
\usepackage{xcolor}
\usepackage{xspace}
\usepackage{enumitem}

\newcommand{\edited}[1]{\textcolor{black}{#1}}

\newcommand{\msft}{Microsoft}
\newcommand{\mlib}{ML.Net\xspace}
\newcommand{\tool}{\textsc{Pretzel}\xspace}
\newcommand{\eat}[1]{}
\newcommand{\stitle}[1]{\vspace{0.6ex}\noindent{\bf #1}}
\newcommand{\at}[1]{\protect\ensuremath{\mathsf{#1}}\xspace}
\newcommand{\etitle}[1]{\vspace{0.8ex}\noindent{\underline{\em #1}}}

\definecolor{dkgreen}{rgb}{0,0.6,0}
\definecolor{gray}{rgb}{0.5,0.5,0.5}
\definecolor{mauve}{rgb}{0.58,0,0.82}

\lstdefinelanguage{Scala}{
  keywords={typeof, new, true, false, catch,def,val, function, return, null, catch, switch, var, if, in, while, do, else, case, break},
  keywordstyle=\color{blue}\bfseries,
  ndkeywords={class, export,extends, boolean, throw, implements, import, this, abstract},
  ndkeywordstyle=\color{dkgreen}\bfseries,
  otherkeywords={+, =>,<=, ==, >,< , ||},
  identifierstyle=\color{black},
  sensitive=false,
  comment=[l]{//},
  morecomment=[s]{/*}{*/},
  commentstyle=\color{purple}\ttfamily,
  stringstyle=\color{red}\ttfamily,
  morestring=[b]',
  morestring=[b]",
  moredelim=**[is][\color{red}]{@}{@},
}
\lstdefinestyle{fault}{ numbers=none, xleftmargin=1.5em , otherkeywords={ =>,<=, ==, > , ||}}

\begin{document}

\title{\tool: Opening the Black Box of Machine Learning \newline Prediction Serving Systems}

\author{
{\rm Yunseong Lee}\\
Seoul National University
\and
{\rm Alberto Scolari}\\
Politecnico di Milano
\and
{\rm Byung-Gon Chun}\\
Seoul National University
\and
{\rm Marco Domenico Santambrogio}\\
Politecnico di Milano
\and
{\rm Markus Weimer}\\
Microsoft
\and
{\rm Matteo Interlandi}\\
Microsoft
}
\date{}
\maketitle

\subsection*{Abstract}
\vspace{-1ex}
Machine Learning models are often composed of pipelines of transformations.
While this design allows to efficiently execute single model components at training-time, prediction serving has different requirements such as low latency, high throughput and graceful performance degradation under heavy load.
Current prediction serving systems consider models as black boxes, whereby prediction-time-specific optimizations are ignored in favor of ease of deployment.
In this paper, we present \tool, a prediction serving system introducing a novel white box architecture enabling both end-to-end and multi-model optimizations. 
Using production-like model pipelines, our experiments show that \tool is able to introduce performance improvements over different dimensions; compared to state-of-the-art approaches \edited{\tool is on average able to reduce 99th percentile latency by 5.5$\times$ while reducing memory footprint by 25$\times$, and increasing throughput by 4.7$\times$.}

\section{Introduction}
\label{sec:intro}
\begin{figure}[h]
\centering
 \includegraphics[width=0.35\textwidth]{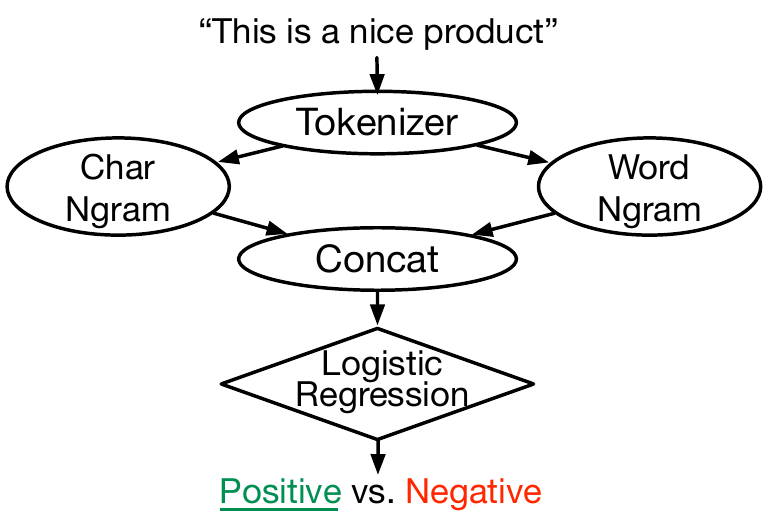}
  	\caption{A Sentiment Analysis (SA) pipeline consisting of operators for featurization (ellipses), followed by a ML model (diamond). \emph{Tokenizer} extracts tokens (e.g., words) from the input string. \emph{Char} and \emph{Word Ngrams} featurize input tokens by extracting n-grams. \emph{Concat} generates a unique feature vector which is then scored by a \emph{Logistic Regression} predictor. This is a simplification: the actual DAG contains about 12 operators.}
    \label{fig:dag}
    \vspace{-4ex}
\end{figure}

\vspace{-2ex}
Many Machine Learning (ML) frameworks such as Google TensorFlow~\cite{tensorflow}, Facebook Caffe2~\cite{caffe2}, Scikit-learn~\cite{scikit}, or \msft~  \mlib~\cite{mldotnet} allow data scientists to declaratively author pipelines of transformations to train models from large-scale input datasets. Model pipelines are internally represented as Directed Acyclic Graphs (DAGs) of operators comprising \emph{data transformations} and \emph{featurizers} (e.g., string tokenization, hashing, etc.), and \emph{ML models} (e.g., decision trees, linear models, SVMs, etc.).
Figure~\ref{fig:dag} shows an example pipeline for text analysis whereby input sentences are classified according to the expressed sentiment.

ML is usually conceptualized as a two-steps process: first, during \emph{training} model parameters are estimated from large datasets by running computationally intensive iterative algorithms; successively, trained pipelines are used for \emph{inference} to generate predictions through the estimated model parameters.
When trained pipelines are served for inference, the full set of operators is deployed altogether. %
 However, pipelines have different system characteristics based on the phase in which they are employed: for instance, at training time ML models run complex algorithms to scale over large datasets (e.g., linear models can use gradient descent in one of its many flavors~\cite{DBLP:journals/corr/Ruder16,NIPS2011_4390,sdca}), while, once trained, they behave as other regular featurizers and data transformations; furthermore, during inference pipelines are often surfaced for direct users' servicing and therefore require low latency, high throughput, and graceful degradation of performance in case of load spikes.

Existing prediction serving systems, such as Clipper~\cite{clipper,clipper2}, TensorFlow Serving~\cite{tf-serving,tf-serving2}, Rafiki~\cite{rafiki}, \mlib~\cite{mldotnet} itself, and others~\cite{predictionIO,redis-ml,tfx,mms} focus mainly on ease of deployment, where pipelines are considered as \emph{black boxes} and deployed into \emph{containers} (e.g., Docker~\cite{docker} in Clipper and Rafiki, \emph{servables} in TensorFlow Serving). 
Under this strategy, only ``pipeline-agnostic'' optimizations such as caching, batching and buffering are available.
Nevertheless, we found that black box approaches fell short on several aspects. 
For instance, prediction services are profitable for ML-as-a-service providers only when pipelines are accessed in batch or frequently enough, and may be not when models are accessed sporadically (e.g., twice a day, a pattern we observed in practice) or not uniformly. 
Also, increasing model density in machines, thus increasing utilization, is not always possible for two reasons: first, higher model density increases the pressure on the memory system, which is sometimes dangerous---we observed (Section~\ref{sec:experiments}) machines swapping or blocking when too many models are loaded; as a second reason, co-location of models may increase tail latency especially when seldom used models are swapped to disk and later re-loaded to serve only a few users' requests.
Interestingly enough, model pipelines often share similar structures and parameters inasmuch as A/B testing and customer personalization are often used in practice in large scale ``intelligent'' services; operators could therefore be shared between ``similar'' pipelines. Sharing among pipelines is further justified by how pipelines are authored in practice: ML pipelines are often produced by fine tuning pre-existing or default pipelines and by editing parameters or adding/removing steps like featurization, etc.

These and other limitations of existing black box systems (further described in Section~\ref{sec:background}) inspired us for developing \tool: a system for serving predictions over trained pipelines originally authored in \mlib and that borrows ideas from the Database and System communities.
Starting from the above observation that trained pipelines often share operators and parameters (such as weights and dictionaries used within operators, and especially during featurization~\cite{Zhang:2016:MOF:2897141.2877204}), we propose a \emph{white box} approach for model serving whereby end-to-end and multi-pipeline optimization techniques are applied to reduce resource utilization while improving performance.
Specifically, in \tool deployment and serving of model pipelines follow a two-phase process. During an \emph{off-line phase}, statistics from training and state-of-the-art techniques from in-memory data-intensive systems~\cite{tupleware,journals/debu/ZukowskiBNH05,conf/cidr/BonczZN05,hyper,Neumann:2011:ECE:2002938.2002940} are used in concert to optimize and compile operators into \emph{model plans}. Model plans are white box representations of input pipelines such that \tool is able to store and re-use parameters  and computation among similar plans.
In the \emph{on-line phase}, memory (data vectors) and CPU (thread-based execution units) resources are pooled among plans. When an inference request for a plan is received, an event-based scheduling~\cite{seda01sosp} is used to bind computation to execution units.

Using \edited{500} different production-like pipelines used internally at \msft, we show the impact of the above design choices with respect to \mlib \edited{and end-to-end solutions such as Clipper}.
\edited{
Specifically, \tool is on average able to improve memory footprint by 25$\times$, reduce the 99th percentile latency by 5.5$\times$, and increase the throughput by 4.7$\times$.
}

In summary, our contributions are:
\vspace{-1ex}
\begin{itemize}
\item A thorough analysis of the problems and limitations burdening black box model serving approaches;
\vspace{-0.5ex}
\item A set of design principles for white box model serving allowing pipelines to be optimized for inference and to share resources;
\vspace{-1ex}
\item A system implementation of the above principles;
\vspace{-1ex}
\item An experimental evaluation showing order-of-magnitude improvements over several dimensions compared to previous black box approaches.
\vspace{-0.5ex}
\end{itemize}

The remainder of the paper is organized as follows: Section~\ref{sec:background} identifies a set of limitations affecting current black box model serving approaches; the outcome of the enumerated limitations is a set of design principles for white box model serving, described in Section~\ref{sec:principles}. Section~\ref{sec:pretzel} introduces the \tool system as an implementation of the above principles. 
Section~\ref{sec:experiments} contains a set of experiments validating the \tool performance, while Section~\ref{sec:limitations} lists the limitations of current \tool implementation and future work. The paper ends with related work and conclusions, respectively in Sections~\ref{sec:related} and~\ref{sec:conclusions}. 

\section{Model Serving: State-of-the-Art and Limitations}
\label{sec:background}
\vspace{-1ex}

Nowadays, ``intelligent'' services such as Microsoft Cortana speech recognition, Netflix movie recommender or Gmail spam detector depend on ML scoring capabilities, which are currently experiencing a growing demand~\cite{Crankshaw:2018:PS:3194653.3210557}. This in turn fosters the research in prediction serving systems in cloud settings~\cite{tf-serving,tf-serving2,clipper,clipper2}, where trained models from data science experts are operationalized. 

Data scientists prefer to use high-level declarative tools such as \mlib, Keras~\cite{keras} or Scikit-learn for better productivity and easy operationalization. These tools provide dozens of pre-defined operators and ML algorithms, which data scientists compose into sequences of operators (called \emph{pipelines}) using %
high-level APIs (e.g., in Python).
\mlib, the ML toolkit used in this paper, is a C\# library that runs on a managed runtime with garbage collection and Just-In-Time (JIT) compilation. Unmanaged C/C++ code can also be employed to speed up processing when possible.
Internally, \mlib operators consume data vectors as input and produce one (or more) vectors as output.~\footnote{Note that this is a simplification. ML.Net in fact support several data types. We refer readers to~\cite{mldotnet2} for more details.}
Vectors are immutable whereby multiple downstream operators can safely consume the same input without triggering any re-execution. Upon pipeline initialization, operators composing the model DAG are analyzed and arranged to form a chain of function calls which, at execution time, are JIT-compiled to form a unique function executing the whole DAG on a single call.
Although \mlib supports Neural Network models, in this work we only focus on pipelines composed by featurizers and classical ML models (e.g., trees, logistic regression, etc.).

\begin{figure}[t]
	\hspace{5ex}\includegraphics[width=.33\textwidth]{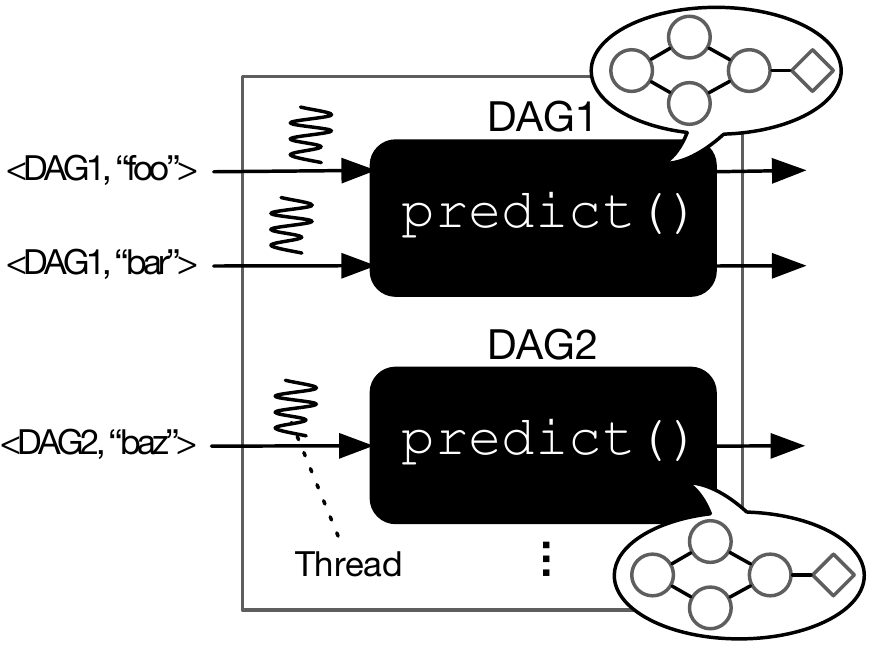}\vspace{1ex}
    \caption{A representation of how existing systems handle prediction requests. Each pipeline is surfaced externally as a black box function. When a prediction request is issued (\texttt{predict()}), a thread is dispatched to execute the chain as a single function call.}
	\label{fig:existing-systems}
        \vspace{-4ex}
\end{figure}

Pipelines are first trained using large datasets to estimate models' parameters. 
\mlib models are exported as compressed files containing several directories, one per pipeline operator, where each directory stores operator parameters in either binary or plain text files.%
~\mlib, as other systems, aims to minimize the overhead of deploying trained pipelines in production by serving them into black box containers, where the same code is used for both training and inference.
Figure~\ref{fig:existing-systems} depicts a set of black box models where the invocation of the function chain (e.g., \texttt{predict()}) on a pipeline returns the result of the prediction: throughout this execution chain, inputs are pulled through each operator to produce intermediate results that are input to the following operators, similarly to the well-known Volcano-style iterator model of databases~\cite{Graefe:1994:VEP:627290.627558}.
To optimize the performance, \mlib (and systems such as Clipper among others) applies techniques such as handling multiple requests in batches and caching the results of the inference if some predictions are frequently issued for the same pipeline. 
However, these techniques assume no knowledge and no control over the pipeline, and are unaware of its internal structure. %
Despite being regarded as a good practice~\cite{google-rules-of-ml}, the black box, container-based design hides the structure of each served model and prevents the system from controlling and optimizing the pipeline execution. 
Therefore, under this approach, there is no principled way neither for sharing optimizations between pipelines, nor to improve the end-to-end execution of individual pipelines.
More concretely, we observed the following limitations in current state-of-the-art prediction serving systems.

\stitle{Memory Waste:} Containerization of pipelines disallows any sharing of resources and runtimes~\footnote{One instance of  model pipeline in production easily occupies 100s of MB of main memory.} between pipelines, therefore only a few (tens of) models can be deployed per machine.
Conversely, ML frameworks such as \mlib have a known set of operators to start with, and \edited{featurizers or} models trained over similar datasets have a high likelihood of sharing parameters. For example, \edited{transfer learning}, A/B testing, and personalized models are common in practice; additionally, tools like \mlib suggest default training configurations to users given a task and a dataset, which leads to many pipelines with similar structure and common objects and parameters. To better illustrate this scenario, we pick a Sentiment Analysis (SA) task with 250 different versions of the pipeline of Figure~\ref{fig:dag} trained by data scientists at \msft. 

\begin{figure}[h]
\vspace{-1ex}
\centering
		\includegraphics[width=.48\textwidth]{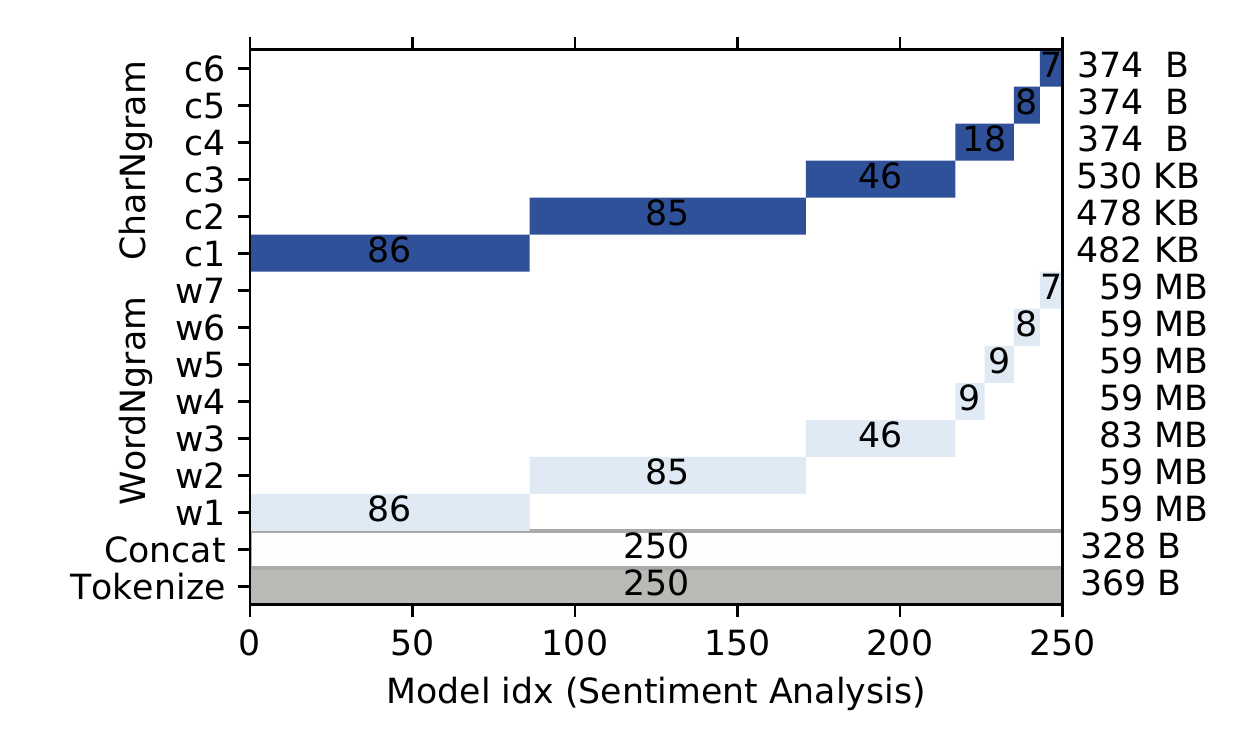}
\vspace{-4ex}
   \caption{\edited{How many identical operators can be shared in multiple SA pipelines. CharNgram and WordNgram operators have variations that are trained on different hyper-parameters. On the right we report operators sizes.}}
	\label{fig:prob-ops-cached}
    \vspace{-2ex}
\end{figure}

Figure~\ref{fig:prob-ops-cached} shows how many different (parameterized) operators are used, and how often they are used within the 250 pipelines. \edited{While some operators like linear regression (whose weights fit in \textasciitilde{}15MB) are unique to each pipeline, and thus not shown in Figure \ref{fig:prob-ops-cached}, many other} operators can be shared among pipelines, therefore allowing more aggressive packing of models: \edited{Tokenize and Concat are used with the same parameters in all pipelines; Ngram operators have only a handful of versions, where most pipelines use the same version of the operators.} This suggests that the resource utilization of current black box approaches can be largely improved. 

\begin{figure}[h]\hspace{1ex}
      \includegraphics[width=0.43\textwidth]{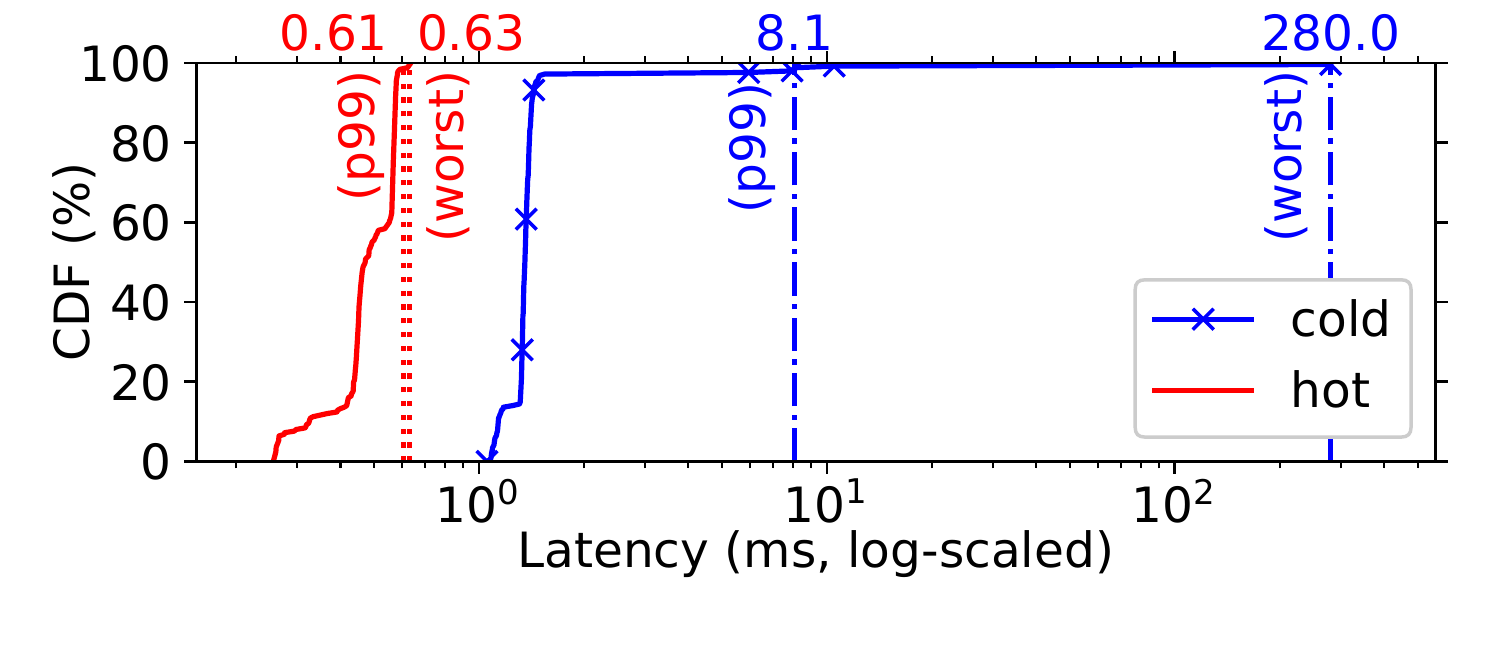}
      \vspace{-2ex}
      \caption{CDF of latency of prediction requests of 250 DAGs. We denote the first prediction as \emph{cold}; the \emph{hot} line is reported as average over 100 predictions after a warm-up period of 10 predictions. \edited{We present the 99th percentile and worst case latency values.}
}
      \label{fig:exp-latency}
          \vspace{-2ex}
\end{figure}

\stitle{Prediction Initialization:} \mlib employs a pull-based execution model that lazily materializes input feature vectors, and tries to reuse existing vectors between intermediate transformations. This largely decreases the memory footprint and the pressure on garbage collection at training time. Conversely, this design forces memory allocation along the data path, thus making latency of predictions sub-optimal and hard to predict.
Furthermore, 
at prediction time \mlib deploys pipelines as in the training phase, which requires initialization of function chain call, reflection for type inference and JIT compilation. While this composability conveniently hides complexities and allows changing implementations during training, it is of little use during inference, when a model has a defined structure and its operators are fixed. In general, the above problems result in difficulties in providing strong tail latency guarantees by ML-as-a-service providers. Figure~\ref{fig:exp-latency} describes this situation, where the performance of \emph{hot} predictions over the 250 sentiment analysis pipelines with memory already allocated and JIT-compiled code is \edited{more than two orders of magnitude faster} than the worst \emph{cold} case version for the same pipelines.

To drill down more into the problem, we found that 57.4\% of the total execution time for a single cold prediction is spent in pipeline analysis and initialization of the function chain, 36.5\% in JIT compilation and the remaining is actual computation time.

\stitle{Infrequent Accesses:}
In order to meet milliseconds-level latencies~\cite{Yun:2015:OAP:2766462.2767708}, model pipelines have to reside in main memory (possibly already warmed-up), since they can have MBs to GBs (compressed) size on disk, with loading and initialization times easily exceeding several seconds. A common practice in production settings is to unload a pipeline if not accessed after a certain period of time (e.g., a few hours). Once evicted, successive accesses will incur a model loading penalty and warming-up, therefore violating Service Level Agreement (SLA).

\begin{figure}[h]
\vspace{-2ex}
	\centering
	\includegraphics[width=0.4\textwidth]{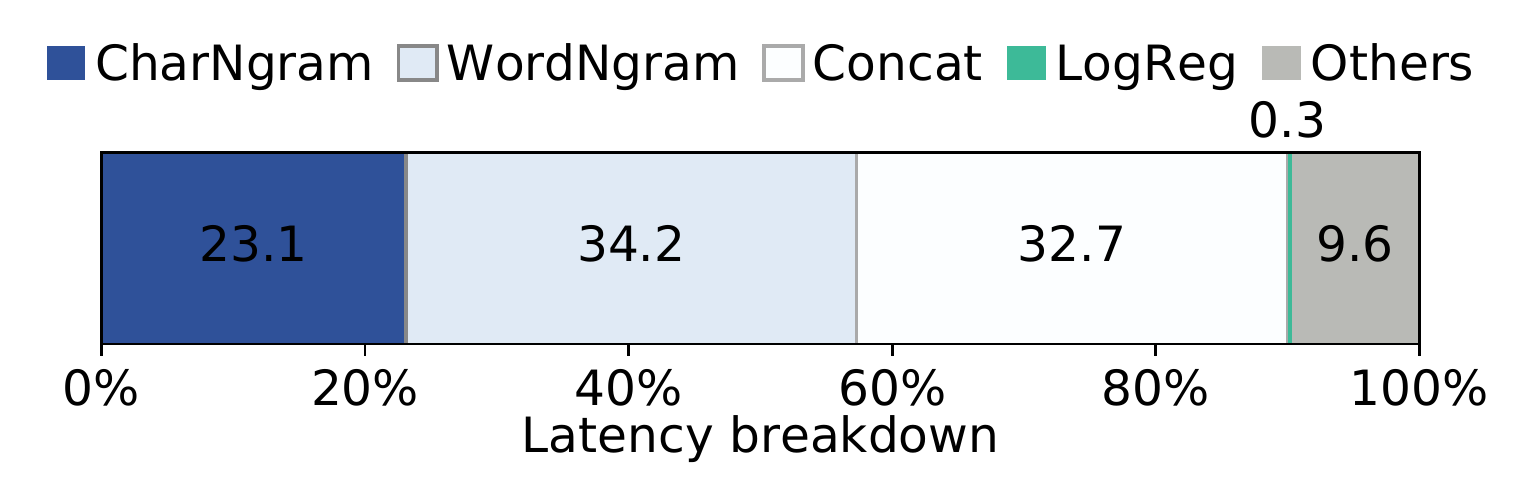}
	\caption{Latency breakdown of a sentiment analysis pipeline: each frame represents the relative wall clock time spent on an operator.}
	\label{fig:execution-breakdown}
        \vspace{-2ex}
\end{figure}

\stitle{Operator-at-a-time Model:}
As previously described, predictions over \mlib pipelines are computed by pulling records through a sequence of operators, each of them operating over the input vector(s) and producing one or more new vectors. While (as is common practice for in-memory data-intensive systems~\cite{Neumann:2011:ECE:2002938.2002940,WandermanMilne2014RuntimeCG,Armbrust:2015:SSR:2723372.2742797}) some interpretation overheads are eliminated via JIT compilation, operators in \mlib (and in other tools) are ``logical'' entities (e.g., linear regression, tokenizer, one-hot encoder, etc.) with diverse performance characteristics. 
Figure~\ref{fig:execution-breakdown} shows the latency breakdown of one execution of the SA pipeline of Figure~\ref{fig:dag}, where the only ML operator (linear regression) takes two orders-of-magnitude less time with respect to the slowest operator (WordNgram).
It is common practice for in-memory data-intensive systems to pipeline operators in order to minimize memory accesses for memory-intensive workloads, and to vectorize compute intensive operators in order to minimize the number of instructions per data item~\cite{tupleware,journals/debu/ZukowskiBNH05}.
\mlib operator-at-a-time model~\cite{journals/debu/ZukowskiBNH05} (as other libraries missing an optimization layer, such as Scikit-learn) is therefore sub-optimal in that computation is organized around logical operators, ignoring how those operators behave together: in the example of the sentiment analysis pipeline at hand, linear regression is commutative and associative (e.g., dot product between vectors) and can be pipelined with Char and WordNgram, eliminating the need for the Concat operation \edited{and the related buffers for intermediate results}. As we will see in the following sections, \tool's optimizer is able to detect this situation and generate an execution plan that is several times faster than the \mlib version of the pipeline.

\stitle{Coarse Grained Scheduling:} Scheduling CPU resources carefully is essential to serve highly concurrent requests and run machines to maximum utilization. Under the black box approach: (1) a thread pool is used to serve multiple concurrent requests to the same model pipeline; 
(2) for each request, one thread handles the execution of a full pipeline sequentially~\footnote{Certain pipelines allow multi-threaded execution, but here we evaluate only single-threaded ones to estimate the per-thread efficiency.}, where one operator is active at each point in time; (3) shared operators/parameters are instantiated and evaluated multiple times (one per container) independently; (4) thread allocation is managed by the OS; and (5) load balancing is achieved ``externally'' by replicating containers when performance degradation is observed.
We found this design sub-optimal, especially in heavily skewed scenarios where a small amount of popular models are scored more frequently then others: indeed, in this setting the popular models will be replicated (linearly increasing the resources used) whereas containers of less popular pipelines will run underutilized, therefore decreasing the total resource utilization.
The above problem is currently out-of-scope for black box, container-based prediction serving systems because they lack visibility into pipelines execution, and they do not allow models to properly share computational resources.

\vspace{2mm}

After highlighting the major inefficiencies of current black box prediction serving systems, we discuss a set of design principles for white box prediction serving. %

\section{White Box Prediction Serving: \\Design Principles}
\label{sec:principles}
\vspace{-1ex}

Based on the observations of Section~\ref{sec:background}, we argue that all previously mentioned limitations can be overcome by embracing a \emph{white box approach} allowing to optimize the execution of predictions both horizontally \emph{end-to-end} and vertically \emph{among multiple model pipelines}.

\stitle{White Box Prediction Serving:} Model containerization disallows any sharing of optimizations, resources, and costs between pipelines. By choosing a white box architecture, pipelines can co-exist on the same runtime; unpopular pipelines can be maintained up and warm, while popular pipelines pay the bills. Thorough scheduling of pipelines' components can be managed within the runtime so that optimal allocation decisions can be made for running machines to high utilization. 
Nevertheless, if a pipeline requires exclusive access to computational or memory resources, a proper reservation-based allocation strategy can be enforced by the scheduler so that container-based execution can be emulated.

\stitle{End-to-end Optimizations:} The operationalization of models for prediction should focus on computation units making optimal decisions on how data are processed and results are computed, to keep low latency and gracefully degrade with load increase. Such computation units should: (1) avoid memory allocation on the data path; (2) avoid creating separate routines per operator when possible, which are sensitive to branch mis-prediction and poor data locality~\cite{Neumann:2011:ECE:2002938.2002940}; and (3) avoid reflection and JIT compilation at prediction time. Optimal computation units can be compiled Ahead-Of-Time (AOT) since pipeline and operator characteristics are known upfront, and often statistics from training are available. The only decision to make at runtime is where to allocate computation units based on available resources and constraints.

\stitle{Multi-model Optimizations:} To take full advantage of the fact that pipelines often use similar operators and parameters (Figure~\ref{fig:prob-ops-cached}), shareable components have to be uniquely stored in memory and reused as much as possible to achieve optimal memory usage. Similarly, execution units should be shared at runtime and resources properly pooled and managed, so that multiple prediction requests can be evaluated concurrently. 
Partial results, for example outputs of featurization steps, can be saved and re-used among multiple similar pipelines.

\section{The Pretzel System}
\label{sec:pretzel}
\vspace{-1ex}

Following the above guidelines, we implemented \tool, a novel white box system for cloud-based inference of model pipelines.
\tool views models as data\-base queries and employs database techniques to optimize DAGs and improve end-to-end performance (Section~\ref{sec:oven}). The problem of optimizing co-located pipelines is casted as a multi-query optimization and techniques such as view materialization (Section~\ref{sec:materialization}) are employed to speed up pipeline execution. Memory and CPU resources are shared in the form of vector and thread pools, such that overheads for instantiating memory and threads are paid upfront at initialization time.

\tool is organized in several components. A \emph{data-flow-style language integrated API} called \at{Flour} (Section~\ref{sec:flour}) with related \emph{compiler} and \emph{optimizer} called \at{Oven} (Section~\ref{sec:oven}) are used in concert to convert \mlib pipelines into \emph{model plans}. An \at{Object} \at{Store} (Section~\ref{sec:store}) saves and shares parameters among plans. A \at{Runtime} (Section~\ref{sec:runtime}) manages compiled plans and their execution, while a \at{Scheduler} (Section~\ref{sec:scheduler}) manages the dynamic decisions on how to schedule plans based on machine workload. Finally, a \at{FrontEnd} is used to submit prediction requests to the system. 

In \tool, deployment and serving of model pipelines follow a two-phase process. During the \emph{off-line phase} (Section~\ref{sec:offline}), \mlib's pre-trained pipelines are translated into \at{Flour} transformations. \at{Oven} optimizer re-arranges and fuses transformations into model plans composed of parameterized logical units called \emph{stages}. Each logical stage is then AOT-compiled into physical computation units where memory resources and threads are pooled at runtime. Model plans are registered for prediction serving in the \at{Runtime} where physical stages and parameters are shared between pipelines with similar model plans. In the \emph{on-line phase} (Section~\ref{sec:online}), when an inference request for a registered model plan is received, physical stages are parameterized dynamically with the proper values maintained in the \at{Object} \at{Store}. The \at{Scheduler} is in charge of binding physical stages to shared execution units.

Figures~\ref{fig:system-design-offline} and~\ref{fig:system-design-online}  pictorially summarize the above descriptions; note that only the on-line phase is executed at inference time, whereas the model plans are generated completely off-line. 
Next, we will describe each layer composing the \tool prediction system.

\subsection{Off-line Phase}
\label{sec:offline}

\subsubsection{Flour}
\label{sec:flour}

The goal of \at{Flour} is to provide an intermediate representation between ML frameworks (currently only \mlib) and \tool, that is both easy to target and amenable to optimizations. Once a pipeline is ported into \at{Flour}, it can be optimized and compiled (Section~\ref{sec:oven}) into a model plan before getting fed into \tool\at{ Runtime} for on-line scoring.
\at{Flour} is a language-integrated API similar to KeystoneML~\cite{keyston-ml}, RDDs~\cite{rdd-nsdi2012} or LINQ~\cite{linq} where sequences of \emph{transformations} are chained into DAGs and lazily compiled for execution. 

Listing~\ref{fig:fcode} shows how the sentiment analysis pipeline of Figure~\ref{fig:dag} can be expressed in \at{Flour}. \at{Flour} programs are composed by transformations where a one-to-many mapping exists between \mlib operators and \at{Flour} transformations (i.e., one operator in \mlib can be mapped to many transformations in \at{Flour}). Each \at{Flour} program starts from a \texttt{FlourContext} object wrapping the \at{Object} \at{Store}. Subsequent method calls define a DAG of transformations, which will end with a call to \texttt{Plan} to instantiate the model plan before feeding it into \tool \at{Runtime}. For example, in lines 2 and 3 of Listing~\ref{fig:fcode} the \texttt{CSV.FromText} call is used to specify that the target DAG accepts as input text in CSV format where \edited{fields are comma separated. %
Line 4 specifies the schema for the input data, where \texttt{TextReview} is a class whose parameters specify the schema fields names, types, and order.
The successive call to \texttt{Select} in line 5 is used to  pick the \texttt{Text} column among all the fields, while the
}
call to \texttt{Tokenize} in line 6 is used to split the input fields into tokens. Lines 8 and 9 contain the two branches defining the char-level and word-level n-gram transformations, which are then merged with the \texttt{Concat} transform in lines 10/11 before the linear binary classifier of line 12. Both char and word n-gram transformations are parameterized by the number of n-grams and maps translating n-grams into numerical format (not shown in the Listing).
Additionally, each \at{Flour} transformation accepts as input an optional set of statistics gathered from training. These statistics are used by the compiler to generate physical plans more efficiently tailored to the model characteristics. Example statistics are max vector size (to define the minimum size of vectors to fetch from the pool at prediction time, as in Section~\ref{sec:online}), dense/sparse representations, etc.

We have instrumented the \mlib library to collect statistics from training and with the related bindings to the \at{Object} \at{Store} and \at{Flour} to automatically extract \at{Flour} programs from pipelines once trained. %
\lstset{numbersep=5pt, xleftmargin=10pt,   framexleftmargin=10pt}
\begin{lstlisting}[label={fig:fcode}, caption={\at{Flour} program for the SA pipeline. Parameters are extracted from the original \mlib pipeline.}]
var fContext = new FlourContext(objectStore, ...) 
var tTokenizer = fContext.CSV
							  .FromText(',')
                       .WithSchema<TextReview>()
                       .Select("Text")
                       .Tokenize();
                       
var tCNgram = tTokenizer.CharNgram(numCNgrms, ...); 
var tWNgram = tTokenizer.WordNgram(numWNgrms, ...);
var fPrgrm = tCNgram
                  .Concat(tWNgram)
                  .ClassifierBinaryLinear(cParams);

return fPrgrm.Plan();
\end{lstlisting}
\vspace{-3ex}
\begin{figure}[t]
\hspace{-2ex}  \includegraphics[width=0.46\textwidth]{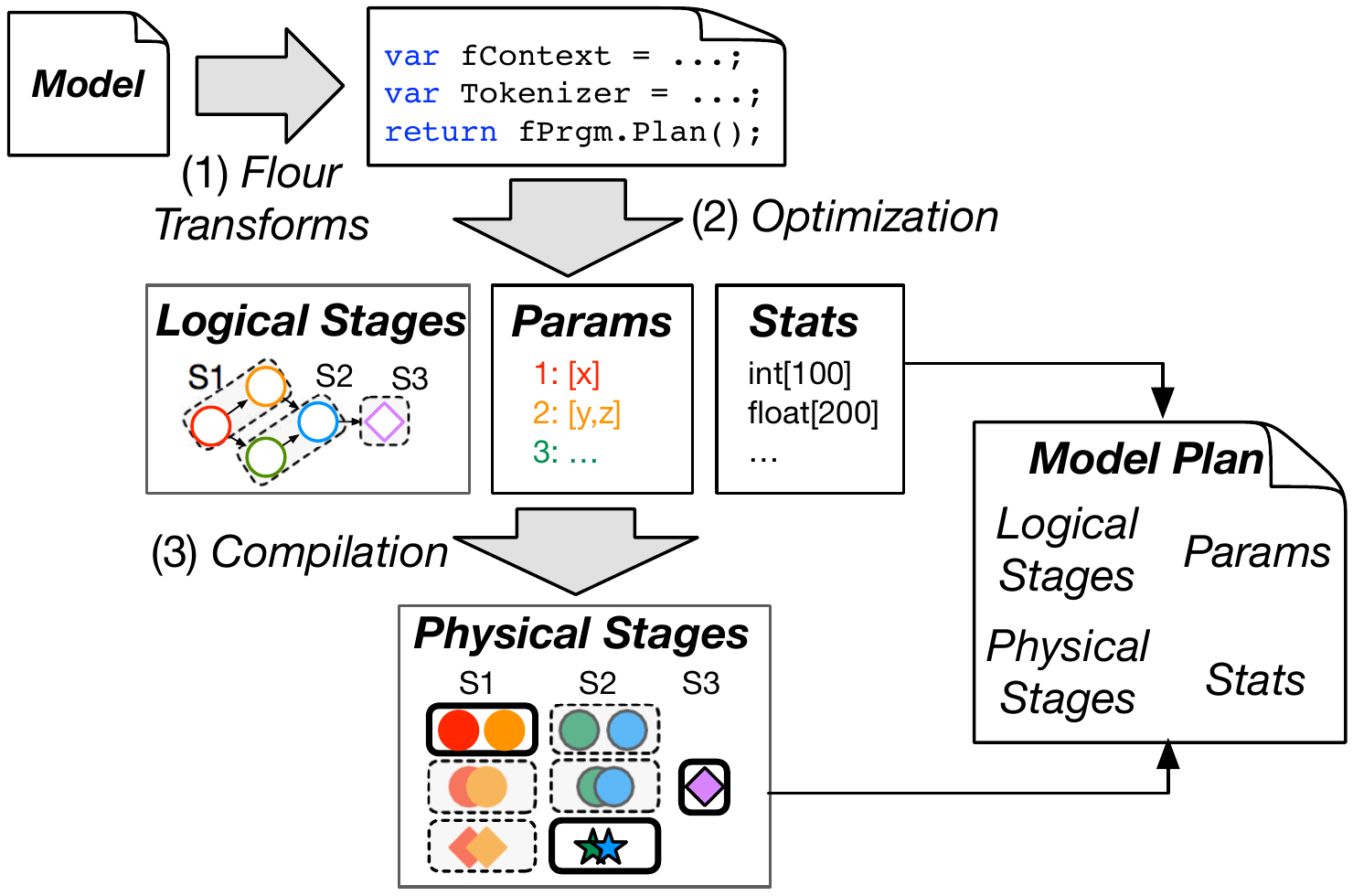}
		\caption{Model optimization and compilation in \tool. In (1), a model is translated into a \at{Flour} program. (2) \at{Oven} \at{ Optimizer} generates a DAG of logical stages from the program. Additionally, parameters and statistics are extracted. (3) A DAG of physical stages is generated by the \at{Oven} \at{Compiler} using logical stages, parameters, and statistics. A model plan is the union of all the elements.}
		\label{fig:system-design-offline}  
            \vspace{-3ex}
\end{figure}
\subsubsection{Oven}
\label{sec:oven}

With \at{Oven}, our goal is to bring query compilation and optimization techniques into \mlib.

\stitle{Optimizer:}
\edited{
When \texttt{Plan} is called on a \at{Flour} transformation's reference (e.g., \texttt{fPrgrm} in line 14 of Listing~\ref{fig:fcode}), all transformations leading to it are wrapped and analyzed.
}%
\at{Oven} follows the typical rule-based database optimizer design where operator graphs (query plans) are transformed by a set of rules until a fix-point is reached (i.e., the graph does not change after the application of any rule). The goal of \at{Oven} \at{Optimizer} is to transform an input graph of \at{Flour} transformations into a stage graph, where each stage contains one or more transformations.
To group transformations into stages we used the Tupleware's hybrid approach~\cite{tupleware}: memory-intensive transformations (such as most featurizers) are pipelined together in a single pass over the data. This strategy achieves best data locality because records are likely to reside in CPU L1 caches~\cite{hyper,Neumann:2011:ECE:2002938.2002940}. Compute-intensive transformations (e.g., vector or matrix multiplications) are executed one-at-a-time so that Single Instruction, Multiple Data (SIMD) vectorization can be exploited, therefore optimizing the number of instructions per record~\cite{journals/debu/ZukowskiBNH05,conf/cidr/BonczZN05}.
Transformation classes are annotated (e.g., 1-to-1, 1-to-n, memory-bound, compute-bound, commutative and associative) to ease the optimization process: no dynamic compilation~\cite{tupleware} is necessary since the set of operators is fixed and manual annotation is sufficient to generate properly optimized plans~\footnote{Note that \mlib does provide a second order operator accepting arbitrary code requiring dynamic compilation. However, this is not supported in our current version of \tool.}.

\edited{
Stages are generated by traversing the \at{Flour} transformations graph repeatedly and applying rules when matching conditions are satisfied.
\at{Oven} \at{Optimizer} consists of an extensible number of \emph{rewriting steps}, each of which in turn is composed of a set of rules performing some modification on the input graph.
Each rewriting step is executed sequentially: within each step, the optimizer iterates over its full set of rules until an iteration exists such that the graph is not modified after all rules are evaluated.
When a rule is active, the graph is traversed (either top-down, or bottom up, based on rule internal behavior; \at{Oven} provides graph traversal utilities for both cases) and the rewriting logic is applied if the matching condition is satisfied over the current node.} 
In its current implementation, the \at{Oven} \at{Optimizer} is composed of 4 rewriting steps:

\etitle{InputGraphValidatorStep:} This step comprises three rules, performing schema propagation, schema validation and graph validation.
Specifically, the rules propagate schema information from the input to the final transformation in the graph, and validate that (1) each transformation's input schema matches with the transformation semantics (e.g., a WordNgram has a string type as input schema, or a linear learner has a vector of floats as input), and (2) the transformation graph is well-formed (e.g., a final predictor exists). 

\etitle{StageGraphBuilderStep:} It contains two rules that rewrite the graph of (now schematized) \at{Flour} transformations into a stage graph.
Starting with a valid transformation graph, the rules in this step traverse the graph until a pipeline-breaking transformation is found, i.e., a Concat or an n-to-1 transformation such as an aggregate used for normalization (e.g., L2).
These transformations, in fact, require data to be fully scanned or materialized in memory before the next transformation can be executed.
For example, operations following a Concat require the full feature vector to be available, or a Normalizer requires the L2 norm of the complete vector.
The output of the \texttt{StageGraphBuilderStep} is therefore a stage graph, where each stage internally contains one or more transformations.
Dependencies between stages are created as aggregation of the dependencies between the internal transformations.
By leveraging the stage graph, \tool is able to considerably decrease the number of vectors (and as a consequence the memory usage) with respect to the operator-at-a-time strategy of \mlib.

\etitle{StageGraphOptimizerStep:}
This step involves 9 rules that rewrite the graph in order to produce an optimal (logical) plan. The most important rules in this step rewrite the stage graph by (1) removing unnecessary branches (similar to common sub-expression elimination); (2) merging stages containing equal transformations (often generated by traversing graphs with branches); (3) inlining stages that contain only one transform; (4) pushing linear models through Concat operations; and (5) removal of unnecessary stages (e.g., when linear models are pushed through Concat operations, the latter stage can be removed if not containing any other additional transformation).

\etitle{OutputGraphValidatorStep:}
This last step is composed of 6 rules. 
These rules are used to generate each stage's schema out of the schemas of the single internal transformations.
Stage schema information will be used at runtime to request properly typed vectors.
Additionally, some training statistics are applied at this step: transformations are labeled as sparse or dense, and dense compute-bound operations are labeled as vectorizable.
A final validation check is run to ensure that the stage graph is well-formed.
\vspace{-2ex}

In the example sentiment analysis pipeline of Figure \ref{fig:dag}, \at{Oven} is able to recognize that the Linear Regression can be pushed into CharNgram and WordNgram, therefore bypassing the execution of Concat. Additionally, Tokenizer can be reused between CharNgram and WordNgram, therefore it will be pipelined with CharNgram (in one stage) and a dependency between CharNgram and WordNgram (in another stage) will be created. The final plan will therefore be composed of 2 stages, versus the initial 4 operators (and vectors) of \mlib.

\stitle{Model Plan Compiler:}
Model plans have two DAGs: a DAG of \emph{logical stages}, and a DAG of \emph{physical stages}. 
Logical stages are an abstraction of the results of the \at{Oven} \at{Optimizer};
physical stages contain the actual code that will be executed by the \tool runtime. For each given DAG, there is a 1-to-n mapping between logical to physical stages so that a logical stage can represent the execution code of different physical implementations. A physical implementation is selected based on the parameters characterizing a logical stage and available statistics.

Plan compilation is a two step process.
After the stage DAG is generated by the \at{Oven} \at{Optimizer}, the Model Plan Compiler (MPC) maps each stage into its logical representation containing all the parameters for the transformations composing the original stage generated by the optimizer.
Parameters are saved for reuse in the \at{Object} \at{Store} (Section~\ref{sec:store}).
Once the logical plan is generated, MPC traverses the DAG in topological order and maps each logical stage into a physical implementation. Physical implementations are AOT-compiled, parameterized, lock-free computation units. 
Each physical stage can be seen as a parametric function which will be dynamically fed at runtime with the proper data vectors and pipeline-specific parameters. This design allows \tool runtime to share the same physical implementation between multiple pipelines and no memory allocation occurs on the prediction path (more details in Section~\ref{sec:runtime}). Logical plans maintain the mapping between the pipeline-specific parameters saved in the \at{Object} \at{Store} and the physical stages executing on the \at{Runtime} \edited{as well as statistics such as maximum vector size (which will be used at runtime to request the proper amount of memory from the pool).} 
Figure~\ref{fig:system-design-offline} summarizes the process of generating model plans out of \mlib pipelines.

\subsubsection{Object Store}
\label{sec:store}

The motivation behind \at{Object} \at{Store} is based on the insights of Figure~\ref{fig:prob-ops-cached}: since many DAGs have similar structures, sharing operators' state (parameters) can considerably improve memory footprint, and consequently the number of predictions served per machine. An example is language dictionaries used for input text featurization, which are often in common among many models and are relatively large. 
The \at{Object} \at{Store} is populated off-line by MPC: when a \at{Flour} program is submitted for planning, new parameters are kept in the \at{Object} \at{Store}, while parameters that already exist are ignored and the stage information is rewritten to reuse the previously loaded one. Parameters equality is computed by looking at the checksum of the serialized version of the objects.

\begin{figure}[t]
	\hspace{2ex}
		\includegraphics[width=0.4\textwidth]{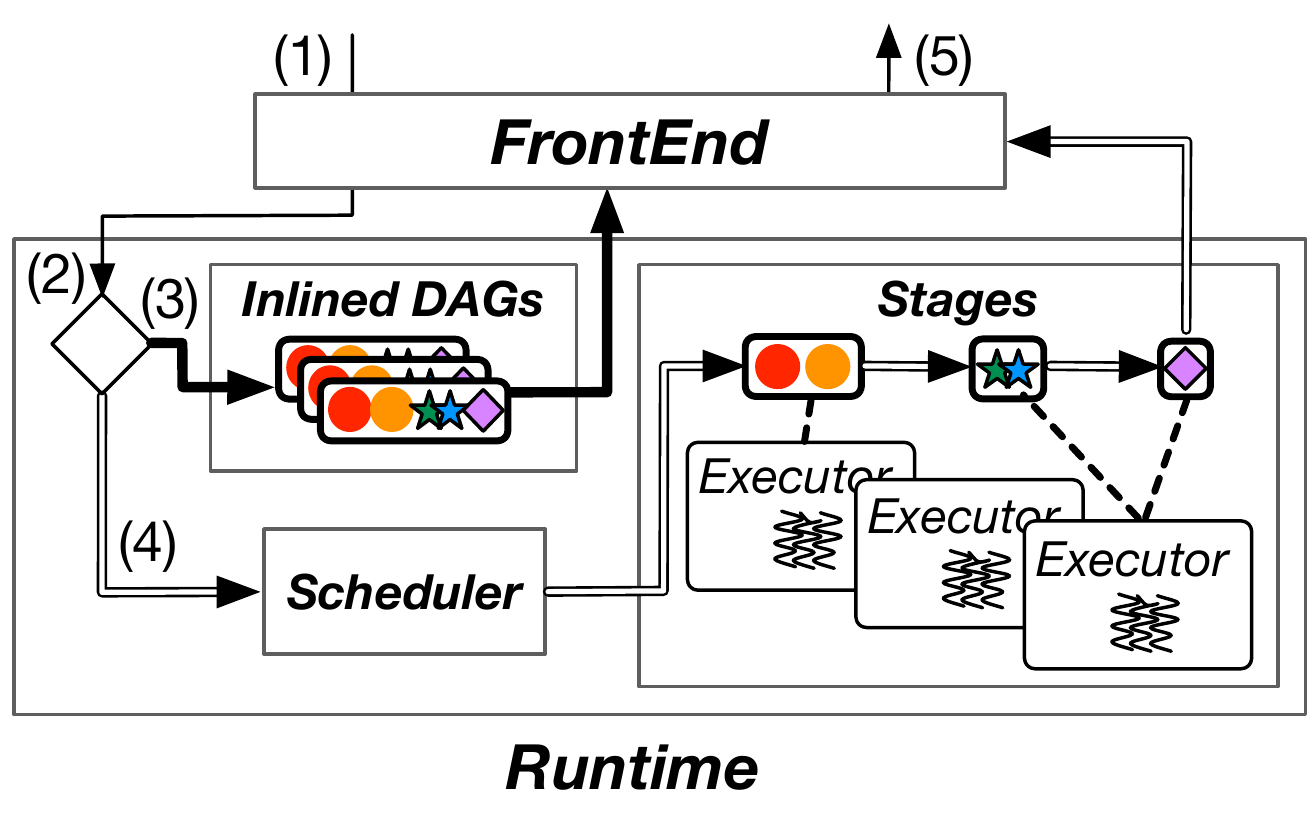}

		\caption{(1) When a prediction request is issued, (2) the \at{Runtime} determines  whether to serve the prediction using (3) the request/response engine or (4) the batch engine. In the latter case, the \at{Scheduler} takes care of properly allocating stages over the Executors running concurrently on CPU cores. (5) The \at{FrontEnd} returns the result to the Client once all stages are complete.}
    \label{fig:system-design-online}
        \vspace{-3ex}
\end{figure}

\subsection{On-line Phase}
\label{sec:online}

\subsubsection{Runtime}
\label{sec:runtime}
\stitle{Initialization:} Model plans generated by MPC are registered in the \tool \at{Runtime}. Upon registration, a unique pipeline ID is generated, and physical stages composing a plan are loaded into a system \emph{catalog}. If two plans use the same physical stage, this is loaded only once in the catalog so that
similar plans may share the same physical stages during execution. 
When the \at{Runtime} starts, a set of vectors and long-running thread pools (called \emph{Executors}) are initialized. Vector pools are allocated per Executor to improve locality~\cite{Gamsa:1999:TML:296806.296814}; Executors are instead managed by the \at{Scheduler} to execute physical stages (Section~\ref{sec:scheduler}) or used to manage incoming prediction requests by the \at{FrontEnd}. Allocations of vector and thread pools are managed by configuration parameters, and allow \tool to decrease the time spent in allocating memory and threads during prediction time.

\stitle{Execution:} 
Inference requests for the pipelines registered into the system can be submitted through the \at{FrontEnd} by specifying the pipeline ID, and a set of input records.
Figure~\ref{fig:system-design-online} depicts the process of on-line inference. 
\tool comes with a \emph{request-response engine} and a \emph{batch engine}. The request-response engine is used by single predictions for which latency is the major concern whereby context-switching and scheduling overheads can be costly. Conversely, the batch engine is used when a request contains a batch of records, or when the prediction time is such that scheduling overheads can be considered as negligible (e.g., few hundreds of microseconds).
The request-response engine inlines the execution of the prediction within the thread handling the request: the pipeline physical plan is JIT-compiled into a unique function call and scored.
Instead, by using the batch engine requests are forwarded to the \at{Scheduler} that decides where to allocate physical stages based on the current runtime and resource status. 
Currently, whether to use the request-response or batch engine is set through a configuration parameter passed when registering a plan. In the future we plan to adaptively switch between the two. 

\subsubsection{Scheduler}
\label{sec:scheduler}
In \tool, model plans share resources, thus scheduling plans appropriately is essential to ensure scalability and optimal machine utilization while guaranteeing the performance requirements.

The \at{Scheduler} coordinates the execution of multiple stages via a late-binding event-based scheduling mechanism similar to task scheduling in distributed systems~\cite{sparrow,rdd-nsdi2012,seda01sosp}: each core runs an Executor instance whereby all Executors pull work from a shared pair of queues: one \emph{low priority} queue for newly submitted plans, and one \emph{high priority} queue for already started stages.
At runtime, a scheduling event is generated for each stage with related set of input/output vectors, and routed over a queue (low priority if the stage is the head of a pipeline, high priority otherwise).
Two queues with different priorities are necessary because of memory requirements. Vectors are in fact requested per pipeline (not per stage) and lazily fulfilled when a pipeline's first stage is being evaluated on an Executor. Vectors are then utilized and not re-added to the pool for the full execution of the pipeline.
Two priority queues allow started pipelines to be scheduled earlier and therefore return memory quickly.

\stitle{Reservation-based Scheduling:}
Upon model plan registration, \tool offers the option to reserve memory or computation resources for exclusive use. Such resources reside on different, pipeline-specific pools, and are not shared among plans, therefore enabling container-like provision of resources. 
Note however that parameters and physical stage objects remain shared between pipelines even if reservation-based scheduling is requested.

\subsection{Additional Optimizations}
\stitle{Sub-plan Materialization:}
\label{sec:materialization}
Similarly to materialized views in database multi-query optimization~\cite{views,views2}, results of installed physical stages can be reused between different model plans. 
When plans are loaded in the runtime, \tool keeps track of physical stages and enables caching of results when a stage with the same parameters is shared by many model plans. 
Hashing of the input is used to decide whether a result is already available for that stage or not.
We implemented a simple Least Recently Used (LRU) strategy on top of the \at{Object} \at{Store} to evict results when a given memory threshold is met.

\stitle{External Optimizations:}
While the techniques described so far focus mostly on improvements that other prediction serving systems are not able to achieve due to their black box nature, \tool \at{FrontEnd} also supports ``external'' optimizations such as the one provided in Clipper and Rafiki.
Specifically, the \at{FrontEnd} currently implements prediction results caching (with LRU eviction policy) and delayed batching whereby inference requests are buffered for a user-specified amount of time and then submitted in batch to the \at{Runtime}. \edited{These external optimizations are orthogonal to \tool's techniques, so both are applicable in a complementary manner.}

\section{Evaluation}
\label{sec:experiments}

\tool implementation is a mix of C\# and C++. %
In its current version, the system comprises 12.6K LOC (11.3K in C\#, 1.3K in C++) and supports about two dozens of \mlib operators, among which linear models (e.g., linear/logistic/Poisson regression), tree-based models, clustering models (e.g., K-Means), Principal Components Analysis (PCA), and several featurizers.

\stitle{Scenarios:} 
The goals of our experimental evaluation are to evaluate how the white box approach performs compared to black box. %
We will use the following scenarios to drive our evaluation: %
\setlist{nolistsep}
\begin{itemize}%
  \item \emph{memory}: in the first scenario, we want to show how much memory saving \tool's
white box approach is able to provide with respect to regular \mlib \edited{and \mlib boxed into Docker containers managed by Clipper.} %
  \item \emph{latency}: this experiment mimics a request/response pattern (e.g.,~\cite{req-res}) such as a personalized web-application requiring minimal latency. \edited{In this scenario, we run two different configurations: (1) a micro-benchmark measuring the time required by a system to render a prediction; and (2) an experiment measuring the total end-to-end latency observed by a client submitting a request.}
  \item \emph{throughput}: this scenario simulates a batch pattern (e.g., ~\cite{batchmsft}) and we use it to assess the throughput of \tool compared to \mlib.
  \item \emph{heavy-load}: we finally mix the above experiments and show \tool's ability to maintain high throughput and graceful degradation of latency, as load increases. To be realistic, in this scenario we generate skewed load across different pipelines. As for the \emph{latency} experiment, we report first the \tool's performance using a micro-benchmark, and then we compare it against the containerized version of \mlib in an end-to-end setting. 
\end{itemize}%

\stitle{Configuration:}
All the experiments reported in the paper were carried out on a  Windows~10 machine with 2 $\times$ 8-core Intel Xeon CPU E5-2620 v4 processors at 2.10GHz with Hyper Threading \edited{disabled}, and 32GB of RAM. 
\edited{
We used .Net Core version 2.0, \mlib version 0.4, and Clipper version 0.2.
For \mlib, we use two black box configurations: a non-containerized one (1 \mlib instance for all models), and a containerized one (1 \mlib instance for each model) where \mlib is deployed as Docker containers running on Windows Subsystem for Linux (WSL) and orchestrated by Clipper.
We commonly label the former as just \mlib; the latter as \mlib + Clipper. 
For \tool we AOT-compile stages using CrossGen~\cite{crossgen}.
For the end-to-end experiments comparing \tool and \mlib + Clipper, we use an ASP.Net \at{FrontEnd} for \tool; the Redis front-end for Clipper.
We run each experiment 3 times and report the median.}

\begin{table}[]
\small
\centering
\caption{\edited{Characteristics  of pipelines in experiments.}}
\label{table:model-info}
\begin{tabular}{@{}lll@{}}
\toprule
Type & \begin{tabular}[]{@{}l@{}}Sentiment\\ Analysis (SA)\end{tabular} & \begin{tabular}[c]{@{}l@{}}\edited{Attendee}\\ \edited{Count (AC)}\end{tabular} \\ \midrule
Input & \begin{tabular}[c]{@{}l@{}}Plain Text \\ (variable length) \end{tabular} & \begin{tabular}[c]{@{}l@{}}Structured Text \\ (40 dimensions)\end{tabular} \\
\hline
Size & \begin{tabular}[c]{@{}l@{}}50MB - 100MB \\ (Mean: 70MB)\end{tabular} & \begin{tabular}[c]{@{}l@{}}10KB - 20MB\\ (Mean: 9MB)\end{tabular} \\
\hline
Featurizers & \begin{tabular}[c]{@{}l@{}}N-gram with \\ dictionaries\\  ($\sim$1M entries)\end{tabular} & \begin{tabular}[c]{@{}l@{}}PCA, KMeans,\\ Ensemble of \\ multiple models\end{tabular} \\
\hline
\end{tabular}
\end{table}
\stitle{Pipelines:}
Table~\ref{table:model-info} describes the two types of model pipelines we use in the experiments: 250 unique versions of Sentiment Analysis (SA) pipeline, and 250 different pipelines implementing \edited{Attendee Count (AC): a regression task used internally to predict how many attendees will join an event}. Pipelines within a category are similar: in particular, pipelines in the SA category benefit from sub-plan materialization, while those in the AC category are more diverse and do not benefit from it.
These latter pipelines comprise several ML models forming an ensemble: in the most complex version, we have a \edited{dimensionality reduction step executed concurrently with a KMeans clustering, a TreeFeaturizer, and multi-class tree-based classifier, all fed into a final tree (or forest) rendering the prediction. SA pipelines are trained and scored over Amazon Review dataset~\cite{He:2016:UDM:2872427.2883037}; AC ones are trained and scored over an internal record of events.}

\subsection{Memory}
\label{sec:memory}

In this experiment, we load all models and report the total memory consumption (model + runtime) per model category. 
SA pipelines are large and therefore we expect memory consumption (and loading time) to improve considerably within this class, proving that \tool's \at{Object} \at{Store} allows to avoid the cost of loading duplicate objects. Less gains are instead expected for the \edited{AC} pipelines because of their small size.

\begin{figure}[h]
\hspace{-2ex}
\includegraphics[width=0.5\textwidth]{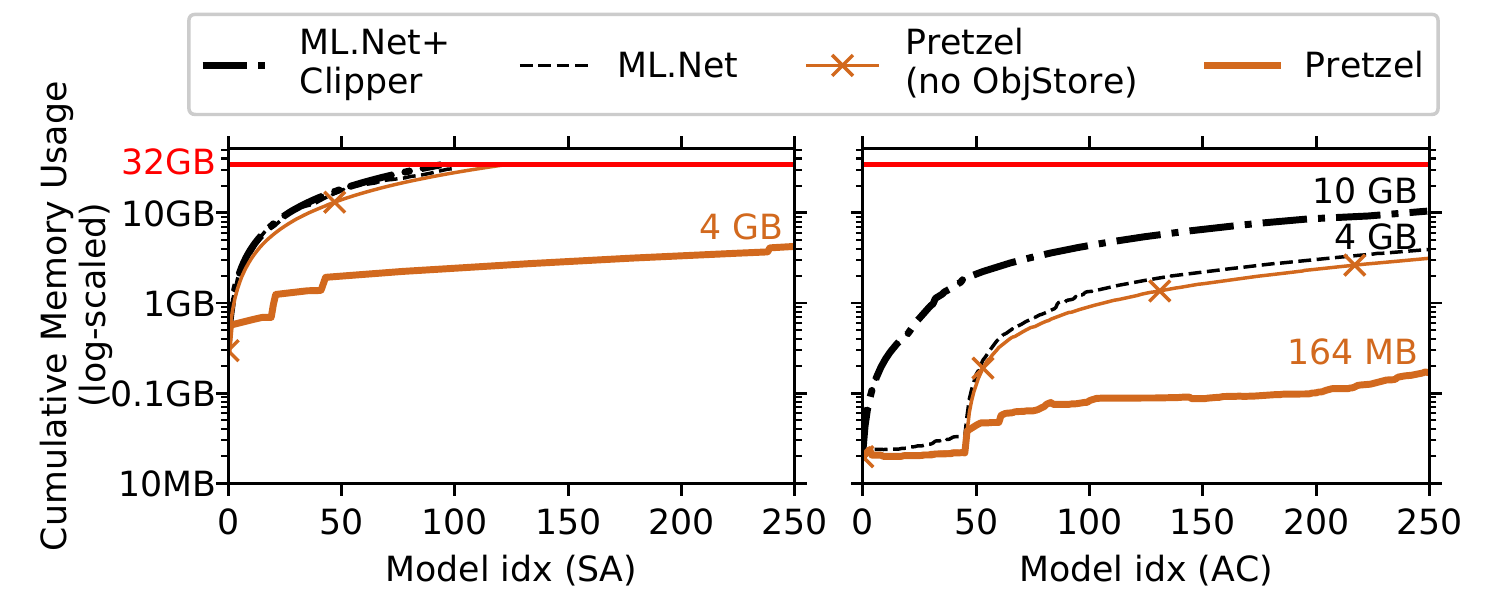}
\vspace{-2ex}  	\caption{\edited{Cumulative memory usage (log-scaled) of the  pipelines in \tool, \mlib and  \mlib + Clipper. The horizontal line represents the machine's physical memory (32GB). Only \tool is able to load all SA pipelines within the memory limit. For AC, \tool uses one order of magnitude less memory than \mlib and \mlib + Clipper. The memory usage of \tool without \at{Object} \at{Store} is almost on par with \mlib.}}
    \vspace{-2ex}
  	\label{fig:memory-usage}
\end{figure}
\edited{Figure~\ref{fig:memory-usage} shows the memory usage for loading all the 250 model pipelines in memory, for both categories. 
For SA, only \tool with \at{Object} \at{Store} enabled can load all pipelines.~\footnote{Note that for \mlib, \mlib + Clipper and \tool without \at{Object} \at{Store} configurations we can load more models and go beyond the 32GB limit. However, models are swapped to disk and the whole system becomes unstable.}
For AC, all configurations are able to load the entire working set, however \tool occupies only 164MBs: about 25$\times$ less memory than \mlib and 62$\times$ less than \mlib + Clipper.
Given the nature of AC models (i.e., small in size), from Figure~\ref{fig:memory-usage} we can additionally notice the overhead (around 2.5$\times$) of using a container-based black box approach vs regular \mlib. 
}

\edited{Keeping track of pipelines' parameters also helps reducing the time to load models: \tool takes around 2.8 seconds to load 250 AC pipelines while \mlib takes around 270 seconds. 
For SA pipelines, \tool takes 37.3 seconds to load all 250 pipelines, while ML.Net fills up the entire memory (32GB) and begins to swap objects after loading 75 pipelines in around 9 minutes.}

\subsection{Latency}
\label{sec:latency}

\edited{
In this experiment we study the latency behavior of \tool in two settings. First, we run a micro-benchmark directly measuring the latency of rendering a prediction in \tool. Additionally, we show how \tool's optimizations can improve the latency. Secondly, we report the end-to-end latency observed by a remote client submitting a request through HTTP.}
\begin{figure}[h]
\vspace{-1.5ex}
\hspace{-2ex}
\includegraphics[width=0.51\textwidth]{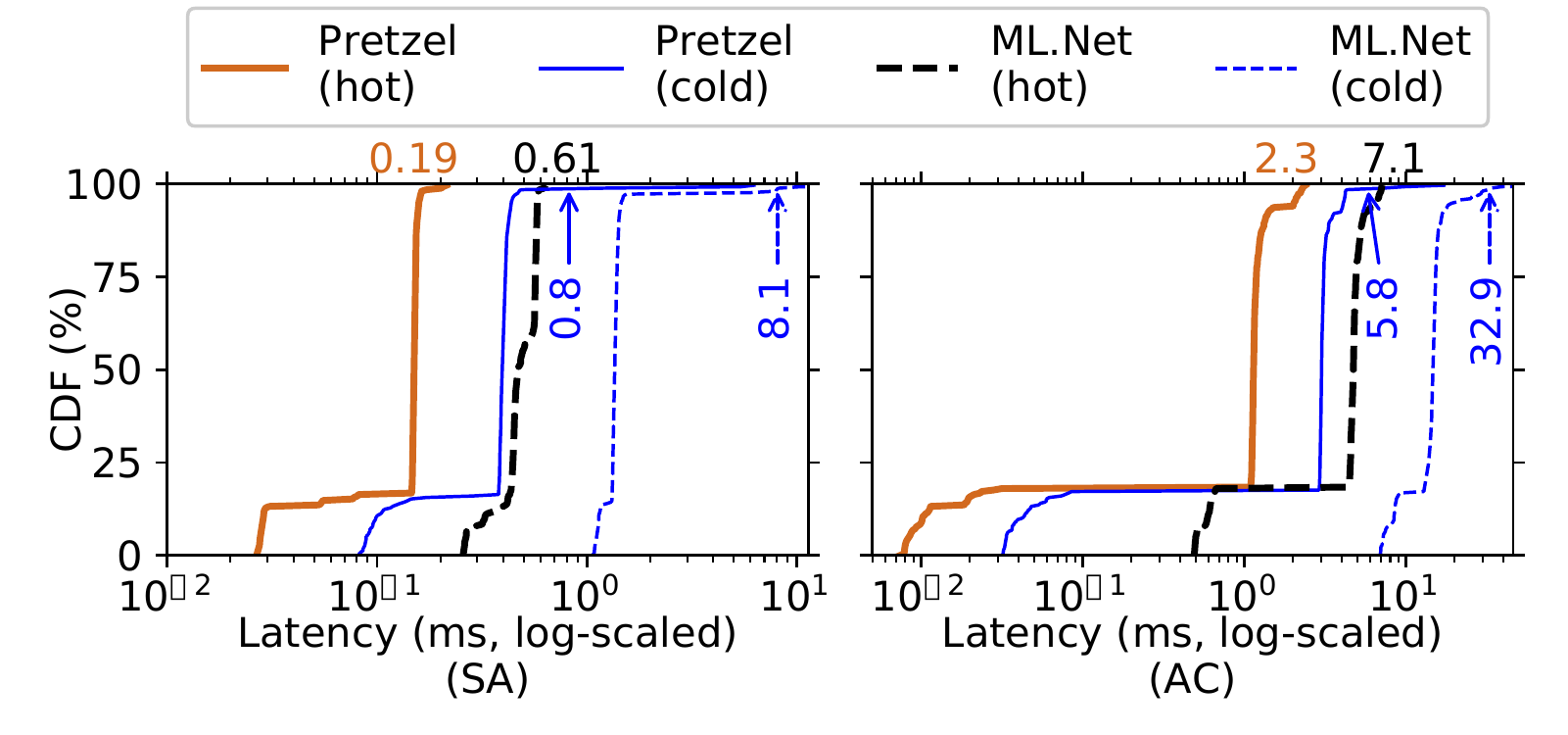}
	\vspace{-2.5ex}
	\caption{Latency comparison between \mlib and \tool. The accompanying blue lines represent the \textit{cold} latency (first execution of the pipelines). On top are the P99 latency values: the hot case is above the horizontal line and the cold case is annotated with an arrow.}
    \label{fig:latency}
    \vspace{-1ex}
\end{figure}

\subsubsection{Micro-benchmark}
Inference requests are submitted sequentially and in isolation for one model at a time. For \tool we use the request-response engine over one single core.
The comparison between \tool and \mlib \edited{for the SA and AC pipelines} is reported in Figure~\ref{fig:latency}.
\edited{We start with studying \textit{hot} and \textit{cold} cases while comparing \tool and \mlib.} Specifically, we label as cold the first prediction requested for a model; the successive 10 predictions are then discarded and we report hot numbers as the average of the following 100 predictions. 

If we directly compare \tool with \mlib, \tool is 3.2$\times$ and 3.1$\times$ faster than \mlib in the 99th percentile latency in hot case (denoted by $P99_{hot}$), and about 9.8$\times$ and 5.7$\times$ in the $P99_{cold}$ case, for SA and AC pipelines, respectively.
If instead we look at the difference between cold and hot cases relative to each system, \tool again provides improvements over \mlib.
The $P99_{cold}$ is about 13.3$\times$ and 4.6$\times$ the $P99_{hot}$ in \mlib, whereas in \tool $P99_{cold}$ is around 4.2$\times$ and 2.5$\times$ from the $P99_{hot}$ case.
Furthermore, \tool is able to mitigate the long tail latency (worst case) of cold scoring. In SA pipelines, the worst case latency is 460.6$\times$ off the $P99_{hot}$ in ML.Net, whereas \tool shows a 33.3$\times$ difference. %
Similarly, in AC pipelines the worst case is 21.2$\times$ $P99_{hot}$ for \mlib, and 7.5$\times$ for \tool. 
To better understand the effect of \tool's optimizations on latency, we turn on and off some optimizations and compare the performance. %

\stitle{AOT compilation:}
This options allows \tool to pre-load all stage code into cache, removing the overhead of JIT compilation in the cold cases. Without AOT compilation, latencies of cold predictions increase on average by 1.6$\times$ and 4.2$\times$ for SA and AC pipelines, respectively. 

\stitle{Vector Pooling:}
By creating pools of pre-allocated vectors, \tool can minimize the overhead of memory allocation at prediction time. When we do not pool vectors, latencies increase in average by 47.1\% for hot and 24.7\% for cold, respectively.

\begin{figure}[h]
\vspace{0ex}
\centering
   \includegraphics[width=0.40\textwidth]{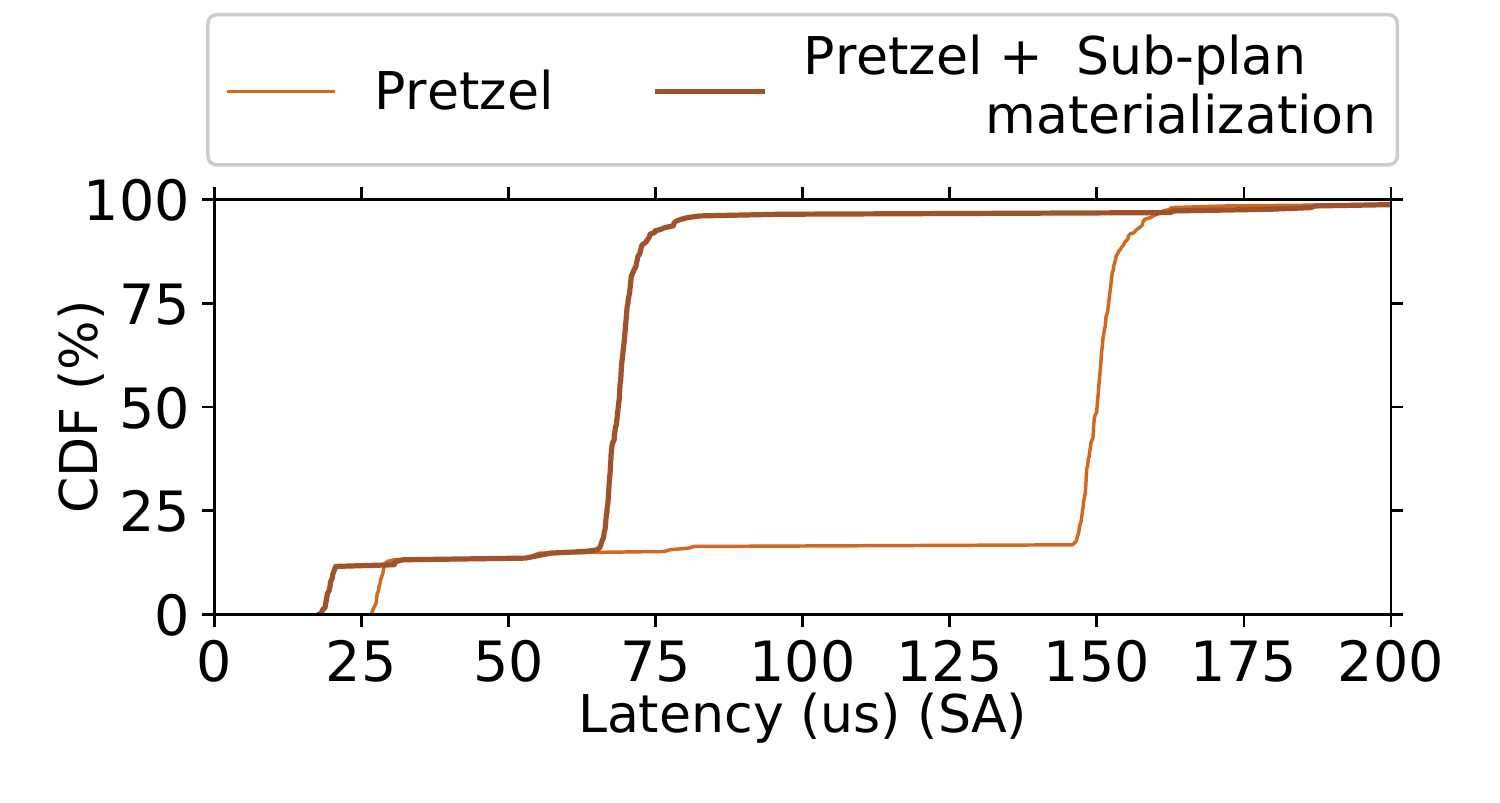}
\vspace{-1ex}
  	\caption{Latency of \tool to run SA models with and without sub-plan materialization. Around 80\% of SA pipelines show more than 2$\times$ speedup. Sub-plan materialization does not apply for AC pipelines.}
  	\label{fig:latency-cdf-w-wo-cache}
        \vspace{-2ex}
\end{figure}

\stitle{Sub-plan Materialization:}
If different pipelines have common featurizers (e.g., SA  as shown in Figure~\ref{fig:prob-ops-cached}),  we can further apply sub-plan materialization to reduce the latency. Figure~\ref{fig:latency-cdf-w-wo-cache} depicts the effect of sub-plan materialization over prediction latency for hot requests. In general, for the SA pipelines in which sub-plan materialization applies, we can see an average improvement of 2.0$\times$, while no pipeline shows performance deterioration.

\begin{figure}
\includegraphics[width=0.47\textwidth]{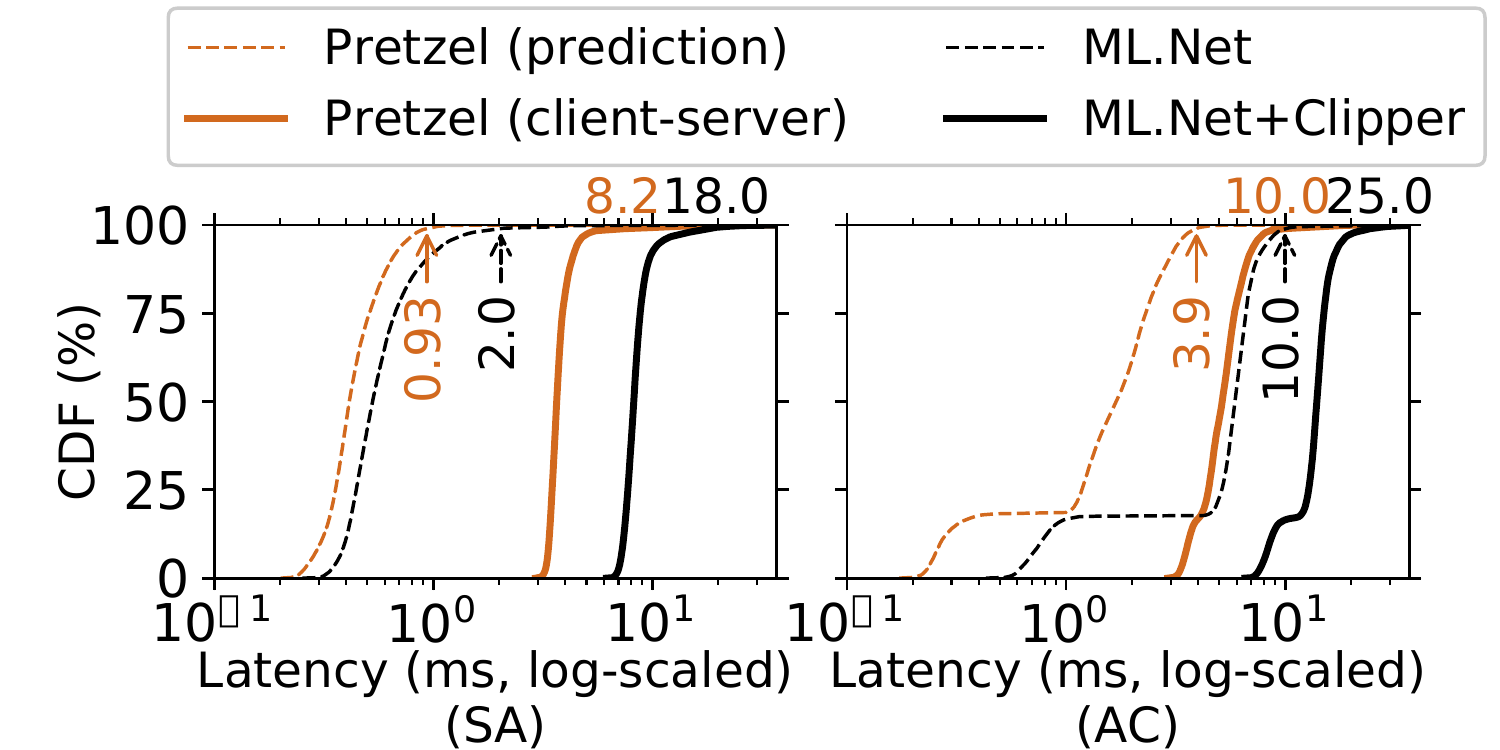}
\vspace{-0ex}
    \caption{
    \edited{The latency comparison between \mlib + Clipper and \tool with ASP.Net \at{FrontEnd}. The overhead of client-server communication compared to the actual prediction is similar in both \tool and \mlib: the end-to-end latency compared to the just prediction latency is 9$\times$ slower in SA and 2.5$\times$ in AC, respectively.}}
    \label{fig:latency-e2e}
        \vspace{-4ex}
\end{figure}

\vspace{-1ex}
\subsubsection{End-to-end}
In this experiment we measure the end-to-end latency from a client submitting a prediction request. For \tool, we use the ASP.Net \at{FrontEnd}, and we compare against \mlib + Clipper. 
The end-to-end latency considers both the prediction latency (i.e., Figure~\ref{fig:latency}) as well as any additional overhead due to client-server communication. 
As shown in Figure~\ref{fig:latency-e2e}, the latter overhead in both \tool and ML.Net + Clipper is in the milliseconds range (around 4ms for the former, and 9 for the latter). Specifically, with \tool, clients observe a latency of 4.3ms at $P99$ for SA models (vs. 0.56ms $P99$ latency of just rendering a prediction) and a latency of 7.3ms for AC models (vs. 3.5ms). 
In contrast, in \mlib+ Clipper, clients observe 9.3ms latency at $P99$ for SA models, and 18.0ms at $P99$ for AC models.
\vspace{-1ex}

\subsection{Throughput}

In this experiment, we run a micro-benchmark assuming a batch scenario where all \edited{500} models are scored several times. We use an API provided by both \tool and \mlib, where we can execute prediction queries in batches: in this experiment we fixed the batch size at 1000 queries. We allocate from 2 up to \edited{13} CPU cores to serve requests, while 3 cores are reserved to generate them. The main goal is to measure the maximum number of requests \tool and \mlib can serve per second. 

\begin{figure}[t]
 \hspace{-1ex}   \includegraphics[width=0.5\textwidth]{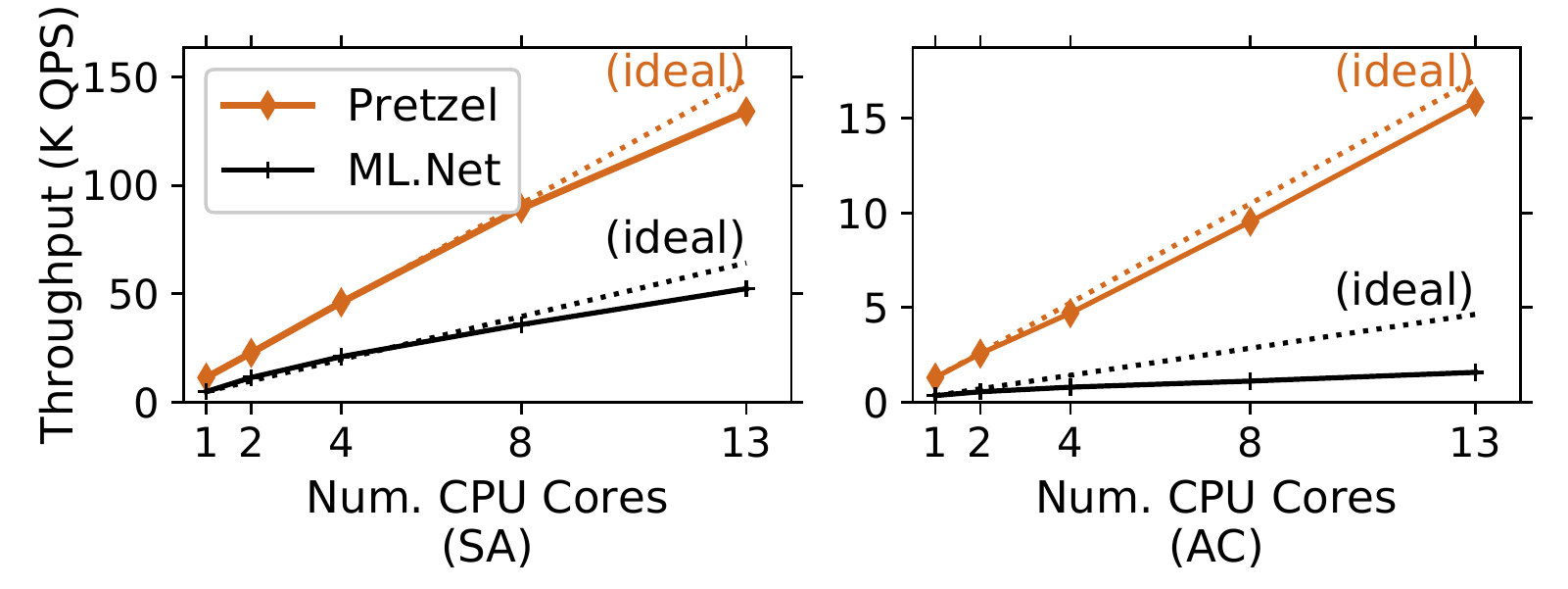}
    \vspace{-3.5ex}
  	\caption{The average throughput computed among the \edited{500} models to process one million inputs each. We scale the number of CPU cores on the x-axis and the number of prediction queries to be served per second on the y-axis. \tool scales linearly to the number of CPU cores.}
  	\label{fig:throughput}
        \vspace{-3ex}
\end{figure}

Figure~\ref{fig:throughput} shows that \tool's throughput (queries per second) is up to \edited{2.6$\times$ higher than \mlib for SA models, 10$\times$ for AC models.
\tool's throughput scales on par with the expected ideal scaling.} 
\edited{Instead, \mlib suffers from higher latency in rendering predictions} and from lower scalability when the number of CPU cores increases. This is because each thread has its own internal copy of models whereby cache lines are not shared, thus increasing the pressure on the memory subsystem: indeed, even if the parameters are the same, the model objects are allocated to different memory areas.

\vspace{-1ex}
\subsection{Heavy Load}
\label{sec:heavy-load}
In this experiment, we show how the performance changes as we change the load. To generate a realistic load, we submit requests to models by following the Zipf distribution ($\alpha=2$).\footnote{The number of requests to the $i$th most popular models is proportional to $i^{-\alpha}$, where $\alpha$ is the parameter of the distribution.} As in Section~\ref{sec:latency}, we first run a micro-benchmark, followed by an end-to-end comparison.

\begin{figure}[h]
\centering
     \includegraphics[width=0.43\textwidth]{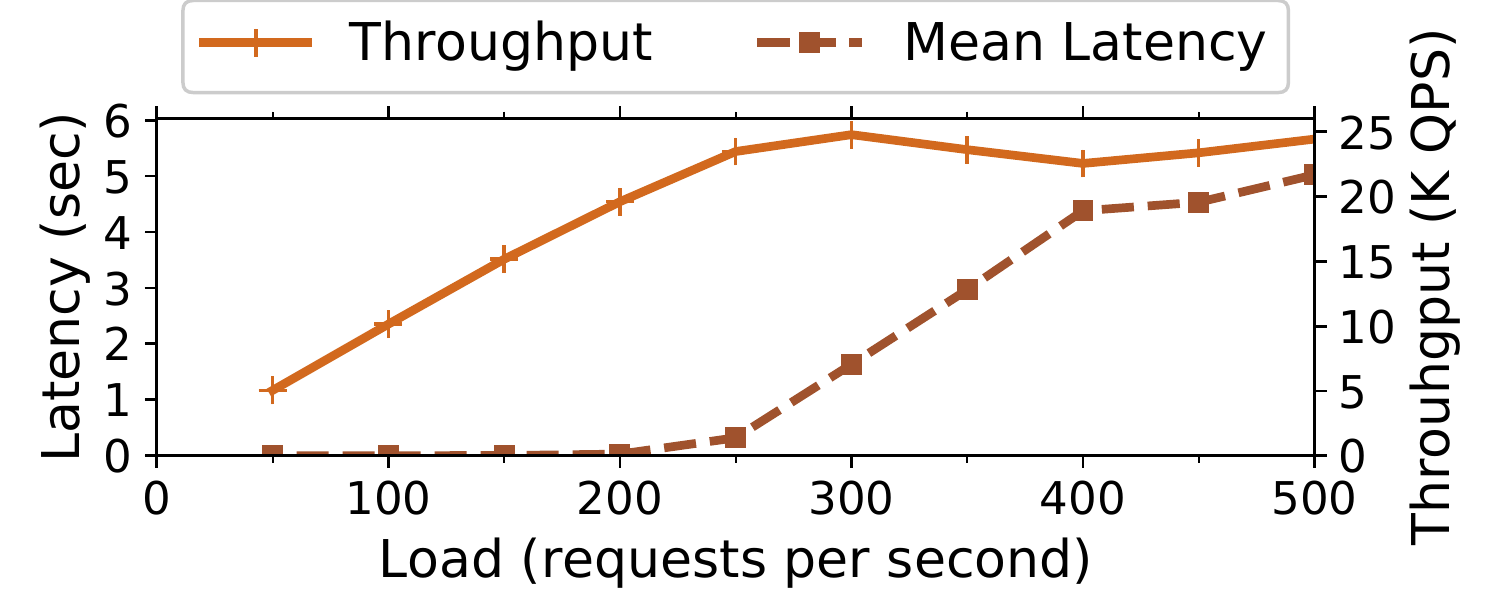}
  	\caption{Throughput and latency of \tool under the heavy load scenario. \edited{We maintain all 500 models in-memory within a \tool instance, and we increase the load by submitting more requests per second. We report latency measurements from latency-sensitive pipelines, and the total system throughput.}}
  	\label{fig:heavy-load}
    \vspace{-3ex}
\end{figure}

\subsubsection{Micro-benchmark}
We load all \edited{500} models in one \tool instance. Among all models, we assume 50\% to be ``latency-sensitive" and therefore we set a batch size of 1. The remaining 50\% models will be requested with 100 queries in a batch.
As in the throughput experiment, we use the batch engine with 13 cores to serve requests and 3 cores to generate load.
Figure~\ref{fig:heavy-load} reports the average latency of latency-sensitive models and the total system throughput under different load configurations. 
As we increase the number of requests, \tool's throughput increases linearly until it stabilizes at about 25k queries per second.
Similarly, the average latency of latency-sensitive pipelines gracefully increases linearly with the load.

\stitle{Reservation Scheduling:}
If we want to guarantee that the performance of latency-critical pipelines is not degrading excessively even under high load, we can enable reservation scheduling. If we run the previous experiment reserving one core (and related vectors) for one model, this does not encounter any degradation in latency (max improvement of 3 orders of magnitude) as the load increases, while maintaining similar system throughput.
\vspace{-2ex}

\subsubsection{End-to-end}
\begin{figure}[t]
	\hspace{-1.2ex}
    \vspace{-1ex}
    \includegraphics[width=0.5\textwidth]{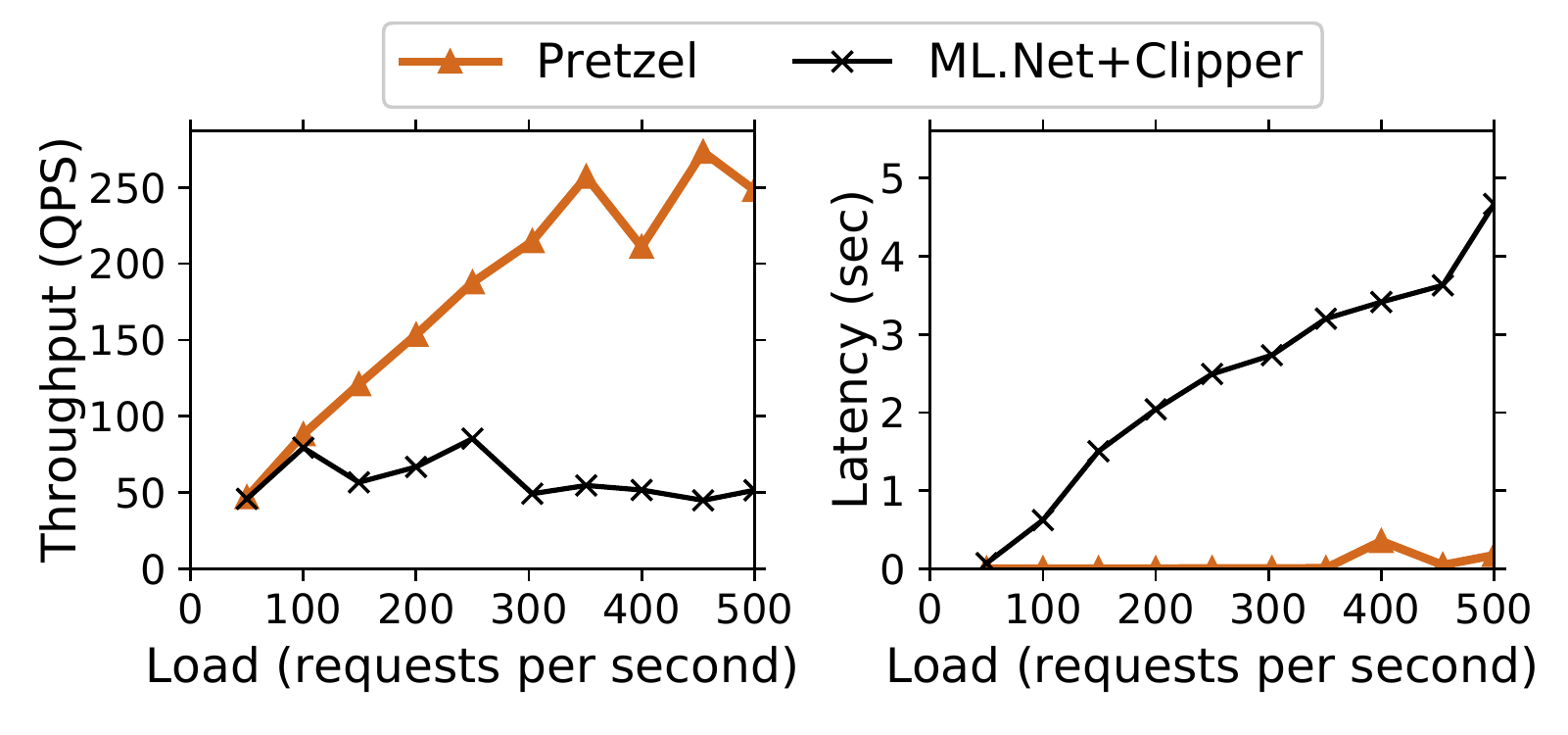}
    \vspace{-2.5ex}
  	\caption{
    \edited{Throughput and latency of \tool and \mlib + Clipper under the end-to-end heavy load scenario. We use 250 AC pipelines to allow both systems to have all pipelines in memory.}
    }
  	\label{fig:heavy-load-e2e}
    \vspace{-3ex}
\end{figure}

In this setup, we periodically send prediction requests to 
\tool with the ASP.Net \at{FrontEnd} and \mlib + Clipper. We assume all pipelines to be latency-sensitive, thus we set a batch of 1 for each request.
As we can see in Figure~\ref{fig:heavy-load-e2e}, \tool's throughput keeps increasing up to around 300 requests per second. If the load exceeds that point, the throughput and the latency begin to fluctuate. On the other hand, the throughput of \mlib + Clipper is considerably lower than \tool's and does not scale as the load increases. Also the latency of \mlib + Clipper is several folds higher than with \tool. The difference is due to the overhead of maintaining hundreds of Docker containers; too many context switches occur across/within containers.

\section{Limitations and Future Work}
\label{sec:limitations}

\stitle{Off-line Phase:}
\tool has two limitations regarding \at{Flour} and \at{Oven} design. First, \tool currently has several logical and physical stages classes, one per possible implementation, which make the system difficult to maintain in the long run. 
Additionally, different back-ends (e.g., \tool currently supports operators implemented in C\# and C++, and experimentally on FPGA~\cite{iccd}) require all specific operator implementations. We are however confident that this limitation will be overcome once code generation of stages will be added (e.g., with hardware-specific templates~\cite{conf/icde/KrikellasVC10}).
Secondly, \at{Flour} and \at{Oven} are currently limited to pipelines authored in ML.Net, and porting models from different frameworks to the white box approach may require non-trivial work.
On the long run our goal is, however, to target unified formats such as ONNX~\cite{onnx}; this will allow us to apply the discussed techniques to models from other ML frameworks as well.

\stitle{On-line Phase:}
\tool's fine-grained, stage-based scheduling may introduce additional overheads in contrasts to coarse-grained whole pipeline scheduling due to additional buffering and context switching. However, such overheads are related to the system load and therefore controllable by the scheduler.
Additionally, we found GC overheads to introduce spikes in latency. Although our implementation tries to minimize the number of objects created at runtime, in practice we found that long tail latencies are common.
On white box architectures, failures happening during the execution of a model may jeopardize the whole system. We are currently working on isolating model failures over the target Executor. 
Finally, \tool runtime currently runs on a single-node. An experimental scheduler adds Non Uniform Memory Access (NUMA) awareness to scheduling policies. We expect this scheduler to bring benefits for models served from large instances (e.g., \cite{numa-aws}). 
We expect in the future to be able to scale the approach over distributed machines, with automatic scale in/out capabilities.

\section{Related Work}
\label{sec:related}
\stitle{Prediction Serving:} As from the Introduction, current ML prediction systems~\cite{clipper,clipper2,tf-serving,tf-serving2,predictionIO,DBLP:conf/cidr/CrankshawBGLZFG15,redis-ml,tfx,rafiki,mms} aim to minimize the cost of deployment and maximize code re-use between training and inference phases~\cite{google-rules-of-ml}. 
Conversely, \tool casts prediction serving as a database problem and applies end-to-end and multi-query optimizations to maximize performance and resource utilization.
Clipper and Rafiki deploy pipelines as Docker containers connected through RPC to a front end. Both systems apply external model-agnostic techniques to achieve better latency, throughput, and accuracy. While we employed similar techniques in the \at{FrontEnd}, in \tool we have not yet explored ``best effort'' techniques such as ensembles, straggler mitigation, and model selection.   
TensorFlow Serving deploys pipelines as \textit{Servable}s, which are units of execution scheduling and version management. One Servable is executed as a black box, although users are allowed to split model pipelines and surface them into different Servables, similarly to \tool's stage-based execution. Such optimization is however not automatic.
LASER~\cite{laser} enables large scale training and inference of logistic regression models, applying specific system optimizations to the problem at hand (i.e., advertising where multiple ad campaigns are run on each user) such as caching of partial results and graceful degradation of accuracy.
Finally, runtimes such as Core ML~\cite{coreml} and Windows ML~\cite{windowsml} provide on-device inference engines and accelerators. To our knowledge, only single operator optimizations are enforced (e.g., using target mathematical libraries or hardware), while neither end-to-end nor multi-model optimizations are used.
\edited{As \tool, TVM~\cite{tvm,tvm2} provides a set of logical operators and related physical implementations, backed by an optimizer based on the Halide language~\cite{halide}.
TVM is specialized on neural network models and does not support featurizers nor ``classical'' models.}

\stitle{Optimization of ML Pipelines:}
There is a recent interest in the ML community in building languages and optimizations to improve the execution of ML workloads~\cite{tvm,dynet,DBLP:journals/corr/ChenLLLWWXXZZ15,xla,noscope}. However, most of them exclusively target Neural Networks and heterogeneous hardware. \edited{Nevertheless, we are investigating the possibility to substitute \at{Flour} with a custom extension of Tensor Comprehension~\cite{tc} to express featurization pipelines. This will enable the support for Neural Network featurizers such as word embeddings, as well as code generation capabilities (for heterogeneous devices).
We are confident that the set of optimizations implemented in \at{Oven} generalizes over different intermediate representations.} 

\edited{Uber's Michelangelo~\cite{michelangelo} has a Scala DSL that can be compiled into bytecode which is then shipped with the whole model as a zip file for prediction. 
Similarly, H2O~\cite{h20} compiles models into Java classes for serving.
This is exactly how \mlib currently works. 
Conversely, similar to database query optimizers, \tool rewrites model pipelines both at the logical and at the physical level.}
KeystoneML~\cite{keyston-ml} provides a high-level API for composing pipelines of operators similarly to \at{Flour}, and also features a query optimizer similar to \at{Oven}, albeit focused on distributed training. KeystoneML's cost-based optimizer selects the best physical implementation based on runtime statistics (gathered via sampling), while no logical level optimizations is provided. Instead, \tool provides end-to-end optimizations by analyzing logical plans~\cite{tupleware,hyper,Neumann:2011:ECE:2002938.2002940,conf/cidr/BonczZN05}, while logical-to-physical mappings are decided based on stage parameters and statistics from training.
\edited{Similarly to the SOFA optimizer~\cite{sofa}, we annotate transformations based on logical characteristics.}
MauveDB~\cite{maueveDB} uses regression and interpolation models as database views and optimizes them as such. MauveDB models are tightly integrated into the database, thus only a limited class of declaratively definable models is efficiently supported.
As \tool, KeystoneML and MauveDB provide sub-plan materialization. 

\stitle{Scheduling:}
Both Clipper~\cite{clipper} and Rafiki~\cite{rafiki} schedule inference requests based on latency targets and provide adaptive algorithms to maximize throughput and accuracy while minimizing stragglers, for which they both use ensemble models.
These techniques are external and orthogonal to the ones provided in \tool.
To our knowledge, no model serving system explored the problem of scheduling requests while sharing resource between models, a problem that \tool addresses with techniques similar to distributed scheduling in cloud computing~\cite{sparrow,Zaharia:2010:DSS:1755913.1755940}.
Scheduling in white box prediction serving share similarities with operators scheduling in stream processing systems~\cite{Babcock:2004:OSD:1037091.1037092,Um:2017:SUI:3124680.3124746} and web services~\cite{seda01sosp}. 

\section{Conclusion}
\label{sec:conclusions}
Inspired by the growth of ML applications and ML-as-a-service platforms, this paper identified how existing systems fall short in key requirements for ML prediction-serving, disregarding the optimization of model execution in favor of ease of deployment. 
Conversely, this work casts the problem of serving inference as a database problem where end-to-end and multi-query optimization strategies are applied to ML pipelines.
To decrease latency, we have developed an optimizer and compiler framework generating efficient model plans end-to-end. 
To decrease memory footprint and increase resource utilization and throughput, we allow pipelines to share parameters and physical operators, and defer the problem of inference execution to a scheduler that allows running multiple predictions concurrently on shared resources.

Experiments with production-like pipelines show the validity of our approach in achieving an optimized execution: \tool delivers order-of-magnitude improvements on previous approaches and over different performance metrics.

\vspace{-1ex}
\subsubsection*{Acknowledgments}
We thank our shepherd Matei Zaharia and the anonymous reviewers for their insightful comments. Yunseong Lee and Byung-Gon Chun were partly supported by the MSIT (Ministry of Science and ICT), Korea, under the SW Starlab support program (IITP-2018-R0126-18-1093) supervised by the IITP (Institute for Information \& communications Technology Promotion), and by the ICT R\&D program of MSIT/IITP (No.2017-0-01772, Development of QA systems for Video Story Understanding to pass the Video Turing Test).

\balance

\bibliographystyle{abbrv}
\small
\bibliography{osdi_2018}

\end{document}